\documentclass[conference]{IEEEtran}
\usepackage{microtype}
\usepackage{booktabs} % for professional tables
\usepackage{epsfig}
\usepackage{amsmath}
\usepackage{amssymb}
\usepackage[numbers]{natbib}
\usepackage{multicol}
\usepackage[bookmarks=true]{hyperref}
\usepackage{graphicx}
\usepackage{algorithm}
\usepackage{algorithmic}
\usepackage{microtype}
\usepackage{multirow}
\usepackage{diagbox}
\usepackage{makecell}
\usepackage{booktabs}
\usepackage{caption}
\usepackage{color}
\usepackage{subcaption}

\ifCLASSINFOpdf
\else
\fi
\begin{document}
\title{Can Pruning Improve Certified Robustness \\ of Neural Networks?}

\author{\IEEEauthorblockN{Zhangheng Li}
\IEEEauthorblockA{UT Austin\\ Austin, Texas, USA}
zoharli@utexas.edu
\and
\IEEEauthorblockN{Tianlong Chen}
\IEEEauthorblockA{UT Austin\\ Austin, Texas, USA}
tianlong.chen@utexas.edu
\and
\IEEEauthorblockN{Linyi Li}
\IEEEauthorblockA{UIUC\\ Urbana, Illinois, USA}
linyi2@illinois.edu
\\
\and
\IEEEauthorblockN{Bo Li}
\IEEEauthorblockA{UIUC\\ Urbana, Illinois, USA}
lbo@illinois.edu
\and
\IEEEauthorblockN{Zhangyang Wang}
\IEEEauthorblockA{UT Austin\\ Austin, Texas, USA}
atlaswang@utexas.edu
}

\maketitle

\begin{abstract}
%\boldmath
With the rapid development of deep learning, the sizes of neural networks become larger and larger so that the training and inference often overwhelm the hardware resources. Given the fact that neural networks are often over-parameterized, one effective way to reduce such computational overhead is neural network pruning, by removing redundant parameters from trained neural networks. It has been recently observed that pruning can not only reduce computational overhead but also can improve empirical robustness of deep neural networks (NNs), potentially owing to removing spurious correlations while preserving the predictive accuracies. This paper for the first time demonstrates that pruning can generally improve certified robustness for ReLU-based NNs under the \textit{complete verification} setting. Using the popular Branch-and-Bound (BaB) framework, we find that pruning can enhance the estimated bound tightness of certified robustness verification, by alleviating linear relaxation and sub-domain split problems. We empirically verify our findings with off-the-shelf pruning methods and further present a new stability-based pruning method tailored for reducing neuron instability, that outperforms existing pruning methods in enhancing certified robustness. Our experiments show that by appropriately pruning an NN, its certified accuracy can be boosted up to \textbf{8.2\%} under standard training, and up to \textbf{24.5\%} under adversarial training on the CIFAR10 dataset. We additionally observe the existence of {\it certified lottery tickets} that can match both standard and certified robust accuracies of the original dense models across different datasets. Our findings offer a new angle to study the intriguing interaction between sparsity and robustness, i.e. interpreting the interaction of sparsity and certified robustness via neuron stability. Codes are available at: \href{https://github.com/VITA-Group/CertifiedPruning}{github.com/VITA-Group/CertifiedPruning}.
\end{abstract}


% \fi
%%\IEEEpeerreviewmaketitle

\section{Introduction}

Neural Network (NN)-based framework is a strong general solution to many problems, yet many of these solutions remain impractical for real-world applications of low fault tolerance. A slight perturbation in the raw input sensory data could completely change the predicting behaviors of the networks. Furthermore, researchers have shown that various kinds of targeted adversarial attacks can easily fool the neural networks \cite{szegedy2013intriguing,goodfellow2014explaining}, which poses threat to many deep learning applications. Fortunately, researchers introduced formal methods to verify neural network behaviors, which made it possible to mathematically derive the prediction bound of a neural network w.r.t. a certain input, and thus evaluate the certified robustness of the neural network. For example, in the image classification task, given an input image with some perturbation, if the lower bound of the output probability of the correct label is higher than the upper bound of probabilities of other incorrect labels, we say the model is certifiably robust w.r.t. this image sample. The goal of neural network verification is to estimate the ground-truth bound as close as possible.  
%The certified robustness verification problem for NNs can be formally described as below \cite{xu2020fast}:
%{\it Given a neural network $f$, an input domain $\mathcal{C}$, and a property $\mathcal{P}$. $\forall x\in \mathcal{C}$, does $f(x)$ satisfy $\mathcal{P}$?}
%Typically, $f$ is a classifier, and $x_i\in\mathcal{X}$ is bounded within an $l_p$ norm ball $\mathcal{C}=\{x |\ ||x-x_0||_p \leq \epsilon\}$. The property $\mathcal{P}$ is a set of desired outputs of the neural network $f$ conditioned on certain input set $\mathcal{X}$. 

In this paper, we are concerned about the certified robustness under the {\it complete verification} setting, where the verifier should output the $exact$ bounds given the input domain $\mathcal{C}$, rather than some relaxation of $\mathcal{C}$, given sufficient time. 
% The bound estimation of certified verification is usually done by propagating bounds through the entire network either in forward or backward manner. 
Despite its theoretical appeal, the complete verification of neural networks is known to be a challenging NP-hard problem \cite{katz2017reluplex,weng2018towards}, mainly due to the non-linear activation functions in neural networks, such as Sigmoid and ReLU. The popular \textit{Branch-and-Bound} (\textbf{BaB}) framework \cite{bunel2017unified} utilized the feature of ReLU activations and adopted the classical divide-and-conquer method to solve the complete verification problem. It branches the bound computation into multiple sub-domains recursively on ReLU nodes and computes the bounds on each sub-domain respectively. The time complexity of this framework is exponential, and typically has pre-set time limit for each sample verification.

%\textcolor{blue}{Shorten the next two paragraphs, move some to related work. Keep the logic chain clear: what was wrong, motivating what's next ... until the chain closes to our work.}

Several verifiers \cite{xu2020fast,wang2021beta} based on the BaB framework were later proposed for efficient complete verification. The core problem addressed in these methods is how to estimate the pre-activation bound (i.e. propagated input bounds of the non-linear activation layers) as tight as possible given limited time. To approach this, they use Linear-Relaxation based Perturbation Analysis (LiRPA) to relax non-linear bound propagation with linear ones and use GPU-accelerated BaB methods to further tighten the estimated bounds as much as possible. However, the estimated bound is still loose, mainly because 
(1)~an efficient linear relaxation of multiple non-linear activation layers is loose both empirically~\cite{salman2019convex} and theoretically~\cite{katz2017reluplex,weng2018towards}, where tightening the relaxation requires an exponential number of linear constraints which is inefficient~\cite{tjandraatmadja2020convex};
and (2)~the BaB framework requires solving an exceedingly large number of sub-domains~(which is exponential in the worst case~\cite{katz2017reluplex}) to provide a tight bound, so in practice we often solve only a part of the sub-domains which yields a loose bound.

% \textcolor{red}{ (1) the linear relaxation of non-linear bound propagation: we refer to the proof procedures of \cite{katz2017reluplex,weng2018towards} for this claim; (2) the limited recursive sub-domain split times: as shown in Theorem 1 from \cite{serra2018bounding}, the number of sub-domains of BaB framework has a non-polynomial upper bound, and we empirically find it is often the case that it is often the case that the sub-domains can't be completely visited during BaB process.}

%\subsection{Our Contribution}

%\textcolor{blue}{Need be more clear: (1) pruning can sometimes improve the robustness accuracy - this has been discovered in prior arts; (2) pruning can be utilized to tighten the bound - our contribution? (3) the above two keypoints together will lead to improved certified robustness, and we empirically show existing pruning schemes can co-achieve both.}
Recent efforts \cite{fu2021drawing, gui2019model, hu2019triple, ye2019adversarial, jordao2021effect, xiao2018training} reveal that proper network pruning can empirically enhance neural network robustness to adversarial attacks. We take one step further to argue that pruning can also be utilized to improve the estimated bound tightness of certified robustness verification, by alleviating linear relaxation and sub-domain split problems. Improving empirical robustness (as a surrogate of ``ground-truth" robustness) and verification tightness can together lead to overall measurable certified robustness, and we find that existing pruning schemes can already co-achieve both. Moreover, inspired by that sparsity can eliminate unstable neurons and improve non-linear neuron stability for verification \cite{xiao2018training}, we present a new stability-based pruning method, that even outperforms existing pruning methods on improving certified robustness. As one last ``hidden gem" finding, we demonstrate the existence of \textit{certified lottery tickets}, that generalizes the lottery ticket hypothesis \cite{frankle2018lottery} to the certified robustness field for the first time: it is defined as those sparse subnetworks found by pruning that can match both the standard and certified accuracies of the original dense models. Our contributions are outlined below:
%\vspace{-0.7em}
% \textcolor{blue}{If your method is simple, then your rationale has to be crispy, impressive and inspiring; PLUS your logic should be clear. Why you argue for sparsity? How it improves bound? Did you just apply pruning methods off-the-shelf, with no second thought? What is the reasoning of choosing pruning algorithms? What is new, what makes your draft not a course project??}Concretely, our contributions are as follows:
\begin{itemize}
    \item For the first time, we demonstrate that pruning can generally improve {\it certified} robustness. We analyze pruning effects from the perspectives of both improving ground-truth robustness of the model and the verification bound tightness, and empirically validate it with extensive pruning methods and training schemes. %\textcolor{blue}{This should be your first impressive highlight -> but nothing impressive now}
    
    \item As pruning can be utilized to improve non-linear neuron stability, we propose a novel regularizer for pruning called {\it NRSLoss} (see Figure \ref{nrsloss}) %\textcolor{blue}{This is a pruning score, not a final loss, right? Calling it loss would mislead readers to think you changed the NN objective} 
    that effectively regularizes the neuron stability and outperforms existing pruning methods in enhancing certified robustness.
%    \item We show structured pruning can effectively accelerate certified verification process. By directly pruning channels, we can obtain essentially smaller networks that can be verified faster on certified verifier for reduced model size and unstable neurons.
    \item Our experiments validate the above proposals by presenting verification results under various settings. For example on the CIFAR10 dataset, under certified training, existing pruning methods as well as our proposed NRSLoss-based pruning boost the certified accuracy for 1.6-7.1\% and 8.2\% respectively; under adversarial training, they boost the certified accuracy for 12.5-24.5\%. 
    \item We additionally observe the existence of {\it certified lottery tickets} that can match both standard and certified robust accuracies of the original dense models, using either the classical iterative magnitude pruning (IMP) \cite{frankle2018lottery} or NRSLoss-based pruning. 
    %that have both significantly better standard and certified accuracy under different pruning methods, among which the IMP \cite{frankle2018lottery} and NRSLoss-based pruning outperform others by improving the standard and certified accuracy for 6-7\% and 7.5-8.5\% respectively under 2/255 perturbation on CIFAR10 dataset. \textcolor{blue}{(This last sentence is important, but incoherently jumps out of nowhere! I would argue this certified LTH phenomenon is too downplayed in current writing - how about we making it an independent section, and a main claim?)}
\end{itemize}

\section{Related Work}
%-------------------------------------------------------------------------
\subsection{Incomplete and complete NN verification}
%%\vspace{-0.5em}
Neural Network verification is a critical issue for developing trustworthy and safe AI. Existing verifiers can be divided into either {\it incomplete} or {\it complete} verifiers. Complete verifiers can typically produce tighter bound than incomplete verifiers, but consume much larger computational resources than incomplete verifiers. Typical incomplete verifiers are based on duality \cite{dvijotham2018dual, raghunathan2018certified} and linear approximations \cite{weng2018towards,wong2018provable, zhang2018efficient}, whereas existing complete verifiers are based on satisfiability modulo theories (SMT) \cite{katz2017reluplex,ehlers2017formal}, mixed integer programming (MIP) \cite{tjeng2017evaluating}, convex hull approximatio \cite{muller2021prima}, or Branch-and-Bound (BaB) \cite{bunel2017unified, xu2020fast, wang2021beta}. 

Traditional complete verifiers such as SMT and MIP are computationally expensive and hard to parallelize. To this end, a series of verifiers based on the BaB framework were recently proposed for efficient and parallelizable complete verification. Auto-LiPRA \cite{xu2020automatic} was an early incomplete verifier and certified trainer, that relaxes ReLU non-linearity with a tight linear relaxation. Following that, Fast-and-Complete algorithm \cite{xu2020fast} proposed to combine auto-LiRPA with BaB for tighter bound estimation and use LP solver for completeness check, which is a GPU-parallelizable complete verification method. Beta-CROWN \cite{wang2021beta} further extended Fast-and-Complete algorithm by replacing the LP completeness check with optimizable constraints based on the Lagrange function, and improved the verification efficiency.

%\vspace{-0.5em}
\subsection{Neural network pruning}
%\vspace{-0.5em}
Pruning removes the redundant structures in NNs and reduces the size of parameter numbers from the computation graph of NNs. It not only constitutes an important class of NN model compression methods but also can act as a regularizer for NN training which can improve the performance w.r.t. original unpruned networks. The pruning process can be conducted at different levels or granularities, such as parameter-level  \cite{frankle2018lottery,zhang2018systematic, ma2020pconv}, filter-level  \cite{luo2018thinet, li2016pruning, roy2020pruning} and layer-wise  \cite{wang2018skipnet, wu2018blockdrop, zhang2019all}. Especially, Lottery Ticket Hypothesis (LTH) \cite{frankle2018lottery} claims the existence of independently trainable sparse subnetworks that can match or even surpass the
performance of dense networks. Such sparse subnetworks called ``winning tickets", can be obtained by simple iterative parameter-level pruning. 

%is a simple and effective parameter-level pruning method, by simply masking out weight parameters with global minimal magnitudes. In this project, we explore the relationship between ReLU stability and Lottery Ticket Hypothesis, and try to derive efficient ways to apply LTH pruning to accelerate the complete NN verification.

\begin{figure*}[!h]
  \centering
  \includegraphics[width=0.8\textwidth]{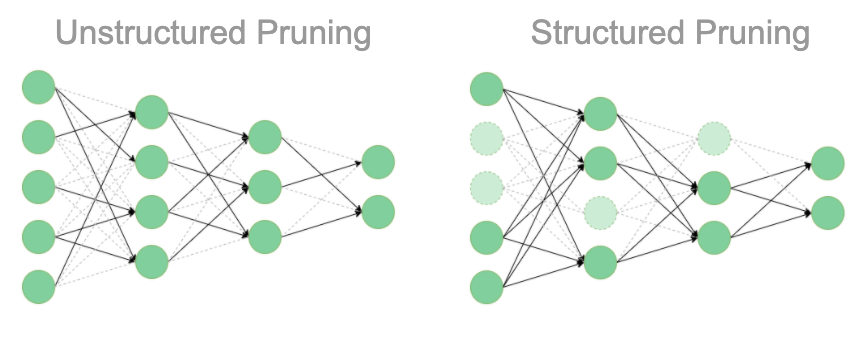}
  \caption{(Figure source from \cite{mark2020what}.) Demonstration of typical unstructured pruning and structured pruning paradigms for a simple Multi-Layer-Perceptron (MLP). Unstructured pruning removes individual weight parameters, while structured pruning removes all input and output weights associated with certain channels.}
  \label{fig_pruning}
\end{figure*}

%\vspace{-0.5em}
\subsection{Pruning and robustness}
%\vspace{-0.5em}
%Intuitively, by pruning the model while preserving the output accuracy, the model might can obtain better robustness since pruning reduces network capacity and provides regularization. 

Recently, several works \cite{ye2019adversarial,gui2019model,jordao2021effect} have revealed that proper network pruning can empirically improve the robustness of a trained NN, potentially due to removing spurious correlations while preserving the predictive accuracies. \cite{fu2021drawing} found that randomly initialized robust subnetworks with better adversarial accuracy than dense model counterparts can be found by IMP. 
%\textcolor{blue}{You missed \cite{xiao2018training} to review here. It should be credited as the first one to inject sparsity to NN weights AND relu stability to speed up verification time. Important for us to clearly make the difference case!} 
\cite{xiao2018training} was the first work to inject sparsity during NN training, with the primary goal to \textit{speed up} certified verification. It considered only weight magnitude pruning, and did not generally demonstrate sparsity to \textit{improve} the achievable certified robustness. %\textcolor{blue}{(\textbf{WRONG comparison. THINK AGAIN!})} however, it only considers single-time pruning with least-weight-magnitude criterion, whereas our work considers different iterative pruning methods with both structured and unstructured sparsity, and reveal general relationship between pruning and certified robustness.
%\textcolor{blue}{The comparison to HYDRA is rather unclear; biggest point is: HYDRA only shows what adv/cert robustness can accomplish under sparsity, but NOT vice versa - what sparsity can do to cert robustness, either accelerate or tighten the bound. }
HYDRA \cite{sehwag2020hydra} incorporated the robustness loss as a pruning objective, and showed such a robustness-aware pruning scheme can lead to high NN sparsity without much robust accuracy loss. Besides studying incomplete certified verification, HYDRA did not specifically analyze what benefits pruning brings for verification, while our NRSLoss-based pruning is explicitly motivated by reducing unstable neurons and tightening the estimated bound in the complete verification.
%\textcolor{blue}{Clarify their motivation of sparsity/pruning is to remove superficial neurons related to patch attacks, hence orthogonal to ours.}
\cite{han2021scalecert} proposed that {\it superficial} neurons that contribute significantly to the feature map values in shallow layers were highly localized and are more prune to{\it adversarial patches}. Hence they used pruning to remove superficial neurons and improved certified defense against adversarial patches -- an orthogonal goal to our work. %\textcolor{blue}{(\textbf{No. EXPAND MORE DETAILS, do not just copy-paste my words!})}

%Despite many progresses on pruning and robustness have been made by prior works, there is no existing work investigating the general effect of network pruning on certified robustness.

\section{Methodology}

\begin{figure*}[h]
\centering
     \begin{subfigure}{.32\linewidth}
         \includegraphics[width=1\linewidth]{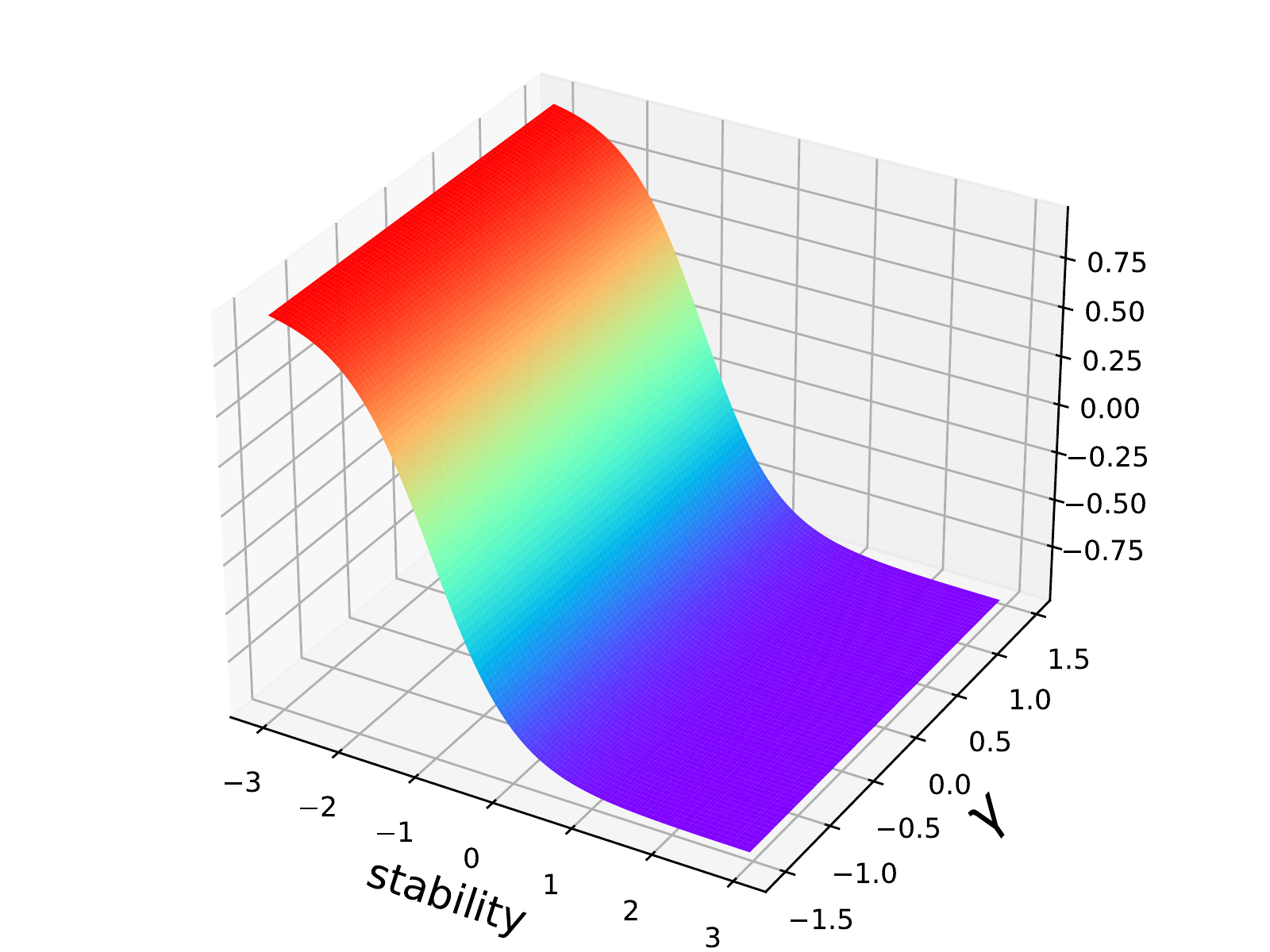}
         \caption{RSLoss}
     \end{subfigure}
     \begin{subfigure}{.32\linewidth}
         \includegraphics[width=1\linewidth]{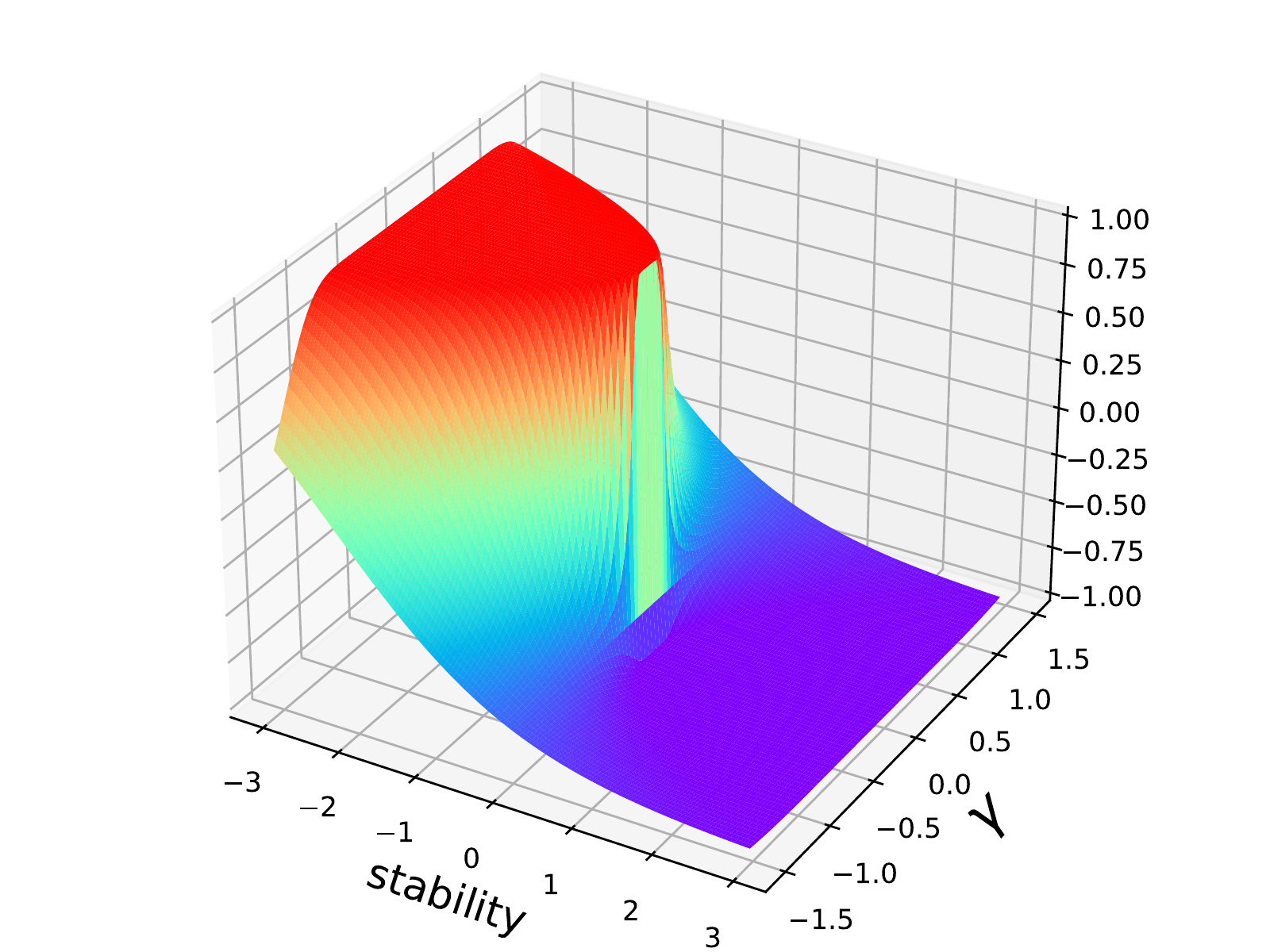}
         \caption{NRSLoss}
     \end{subfigure}
     \begin{subfigure}{.32\linewidth}
         \includegraphics[width=1\linewidth]{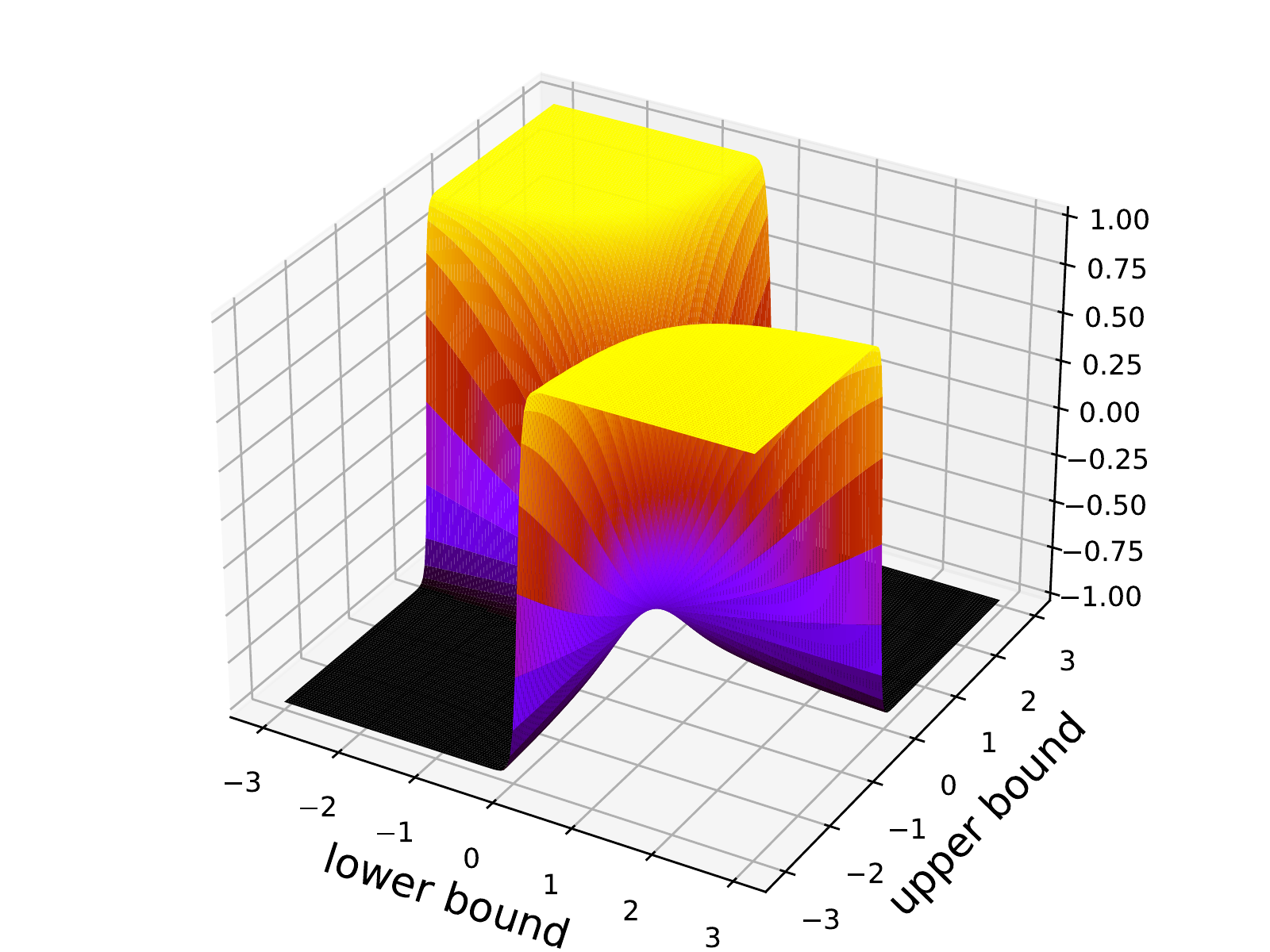}
         \caption{NRSLoss($\gamma=0.5$)}
     \end{subfigure}
     
     \begin{subfigure}{.32\linewidth}
         \includegraphics[width=1\linewidth]{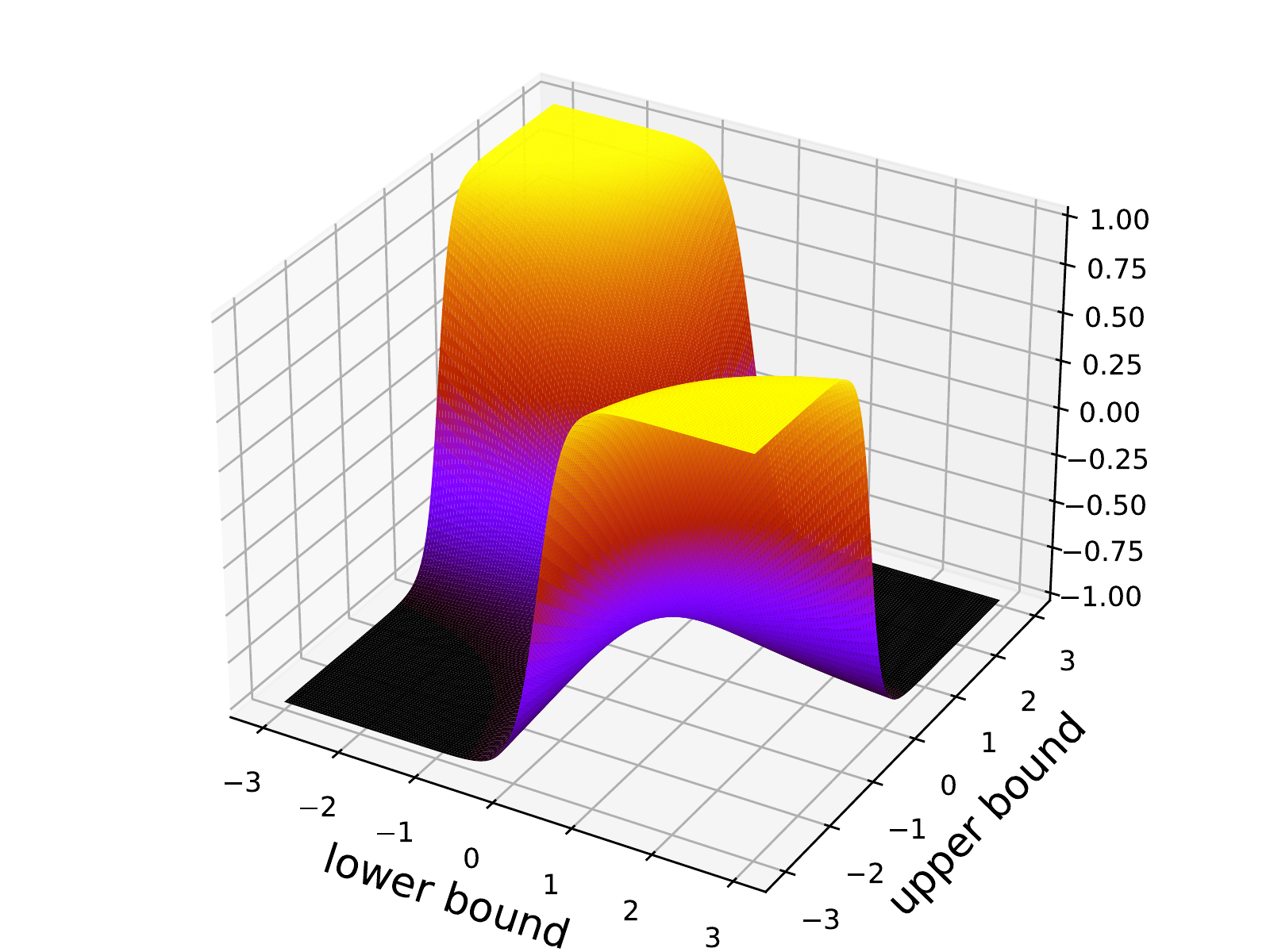}
         \caption{NRSLoss($\gamma=1$) (i.e. RSLoss)}
     \end{subfigure}
     \begin{subfigure}{.32\linewidth}
         \includegraphics[width=1\linewidth]{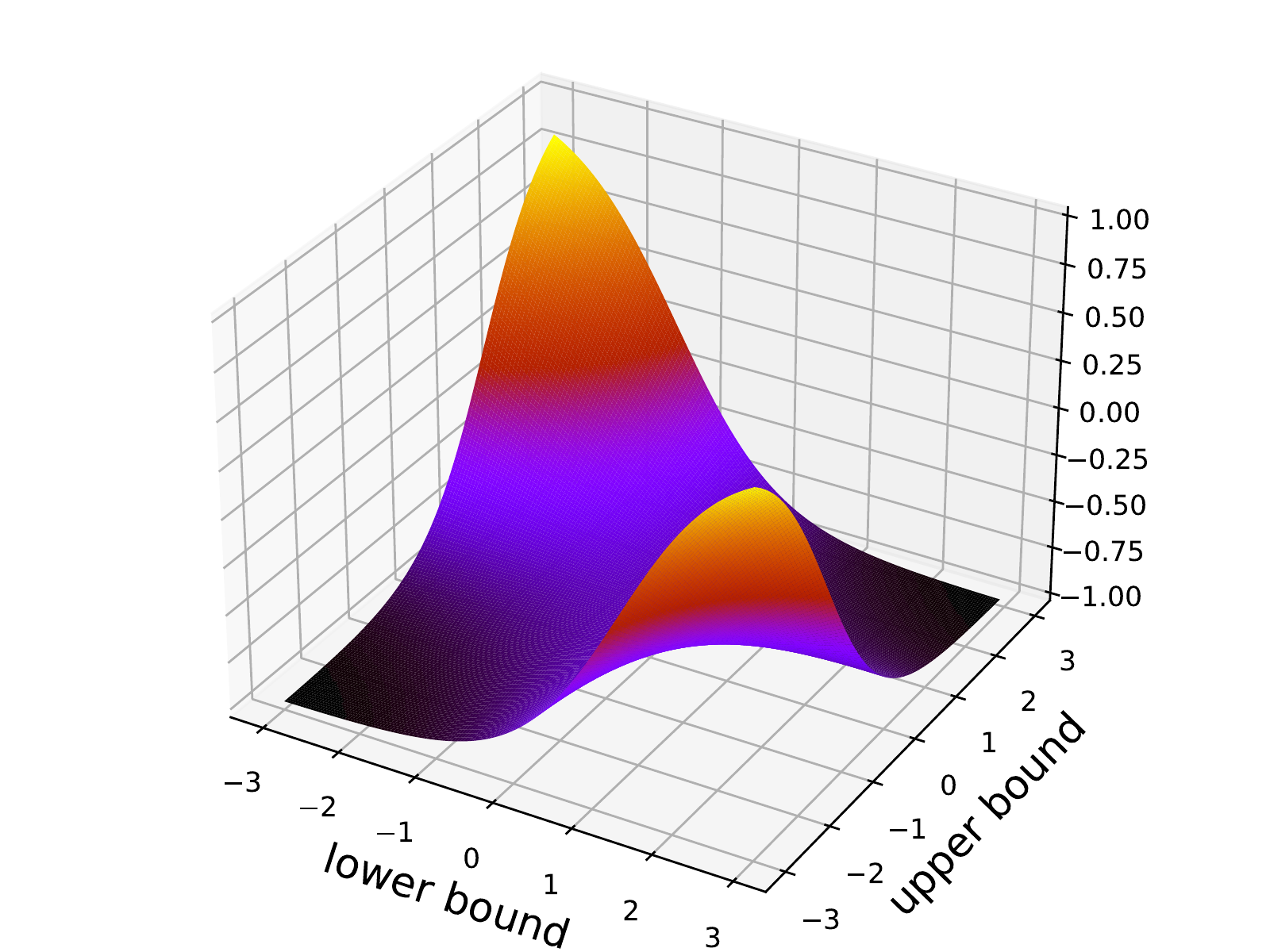}
         \caption{NRSLoss($\gamma=2$)}
     \end{subfigure}
     \begin{subfigure}{.32\linewidth}
         \includegraphics[width=1\linewidth]{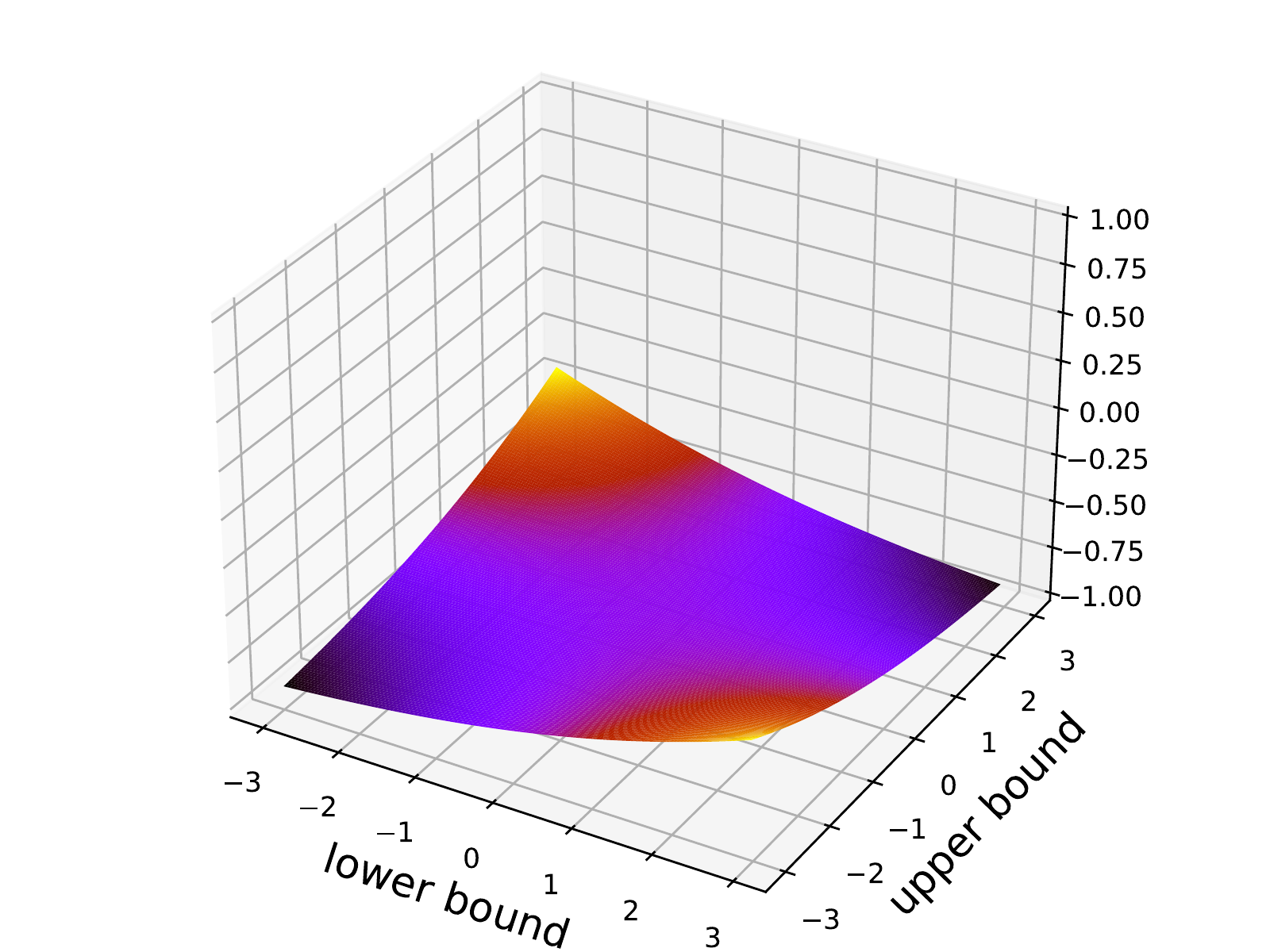}
         \caption{NRSLoss($\gamma=4$)}
     \end{subfigure}
         \caption{(a)(b) The landscape of RSLoss and NRSLoss with varied stability and BN channel weight $\gamma$. {\it Stability} means the stability of a ReLU neuron, i.e. the pre-activation lower bound times upper bound. $\gamma$ is the corresponding channel weight of Batch-Normalization layer, whose magnitude denotes the importance of each channel. The NRSLoss is high when the neuron is unstable and the corresponding channel has low importance. (c)-(f) The sample landscape of NRSLoss with varied lower and upper pre-activation bounds given different fixed $\gamma$. With the growth of $\gamma$ that implies channel(neuron) importance, the NRSLoss gets increasingly suppressed.}
    \label{nrsloss}
\end{figure*}

In this paper, following prior works on certified robustness, we focus on non-linear feed-forward neural networks with ReLU activations. In this section, we first analyze what benefits would network pruning brings to certified robustness and verification, and then introduce the specific pruning methods we test in our experiments.

\subsection{Preliminary}

\subsubsection{Unstructured and Structured Pruning}
Network pruning is one of the most effective model compression paradigms for deep neural network, by removing redundant parameters or neurons from over-parameterized neural networks, and many of them focus on pruning learnable weight parameters. Existing weight pruning methods can be divided to unstructured pruning and structured pruning, depending on whether weights are pruned individually or by group. An example of unstructured pruning v.s. structured pruning for a simple network composed of multiple fully-connected layers is shown in Figure \ref{fig_pruning}. For unstructured pruning, individual weights that connect two channels (neurons) of adjacent layers are removed; for structured pruning, all input and output weights associated with certain channels are removed. Another typical case is pruning for convolutional neural networks, unstructured pruning usually removes individual weight elements of the convolutional kernels, whereas structured pruning removes all input and output kernels associated with certain channels. 

Each of these two pruning paradigms has its advantages and disadvantages over the other. Unstructured pruning can better preserve the performance of the original dense networks due to pruning flexibility on individual weights, but is hard to realize real hardware acceleration during inference. In contrast, the hardware compression during inference for structured pruning can be easily implemented due to the removal of entire channels, but has less pruning flexibility compared to unstructured pruning, which would generally lead to worse performance that unstructured pruning. In this paper, we investigate the influence of both unstructured and structured pruning on certified robustness and are interested in both the performance gain and the reduction of computational overhead brought by pruning.

\subsubsection {Lottery Ticket Hypothesis(LTH)}

The lottery ticket hypothesis \cite{frankle2018lottery} states that a randomly initialized dense neural network contains at least one subnetwork (i.e. by pruning the parameters of the dense network) that has the same initialization of the unpruned parameters and can match the test performance of the dense network after training for at most the same iterations as the dense network, and such subnetworks are called the winning tickets of the dense network. To find these winning tickets, they propose \textbf{Iterative Magnitude Pruning}(IMP) algorithm: Firstly, start from a dense initialization $W_0$, and then train the network until convergence to weight $W_t$. Then we determine the $\rho$ percent smallest magnitude weights in $|W_t|$ and create a binary mask $m_0$ that prunes these. Then retrain the pruned network from the same initialization weight $W_0 \odot m_0$ to convergence. Iterating this procedure will produce subnetworks with different sparsity, among which certain subnetworks can match the test performance of the original dense network, i.e. the winning tickets.

\subsubsection{ReLU Neuron Stability}

\label{unstable}
\begin{figure*}[!h]
  \centering
  \includegraphics[width=\textwidth]{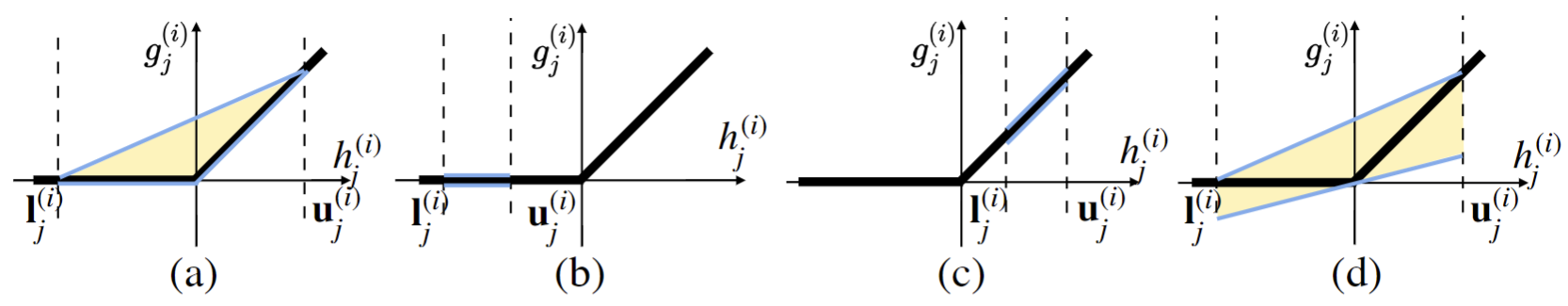}
  \caption{(Figure source from \cite{xu2020fast})) Illustration of the stability of ReLU neuron and linear relaxation.}
  \label{fig_unstable}
\end{figure*}

The illustration of ReLU neuron stability is demonstrated in Figure \ref{fig_unstable}. The ReLU activation function is zero when input value is less than 0, and an identity function when input value is no less than 0. As shown in the Figure \ref{fig_unstable}, ${\bf h}_j^{(i)}$ means the pre-activation value of $j$th ReLU neuron at $i$th layer of the network. ${\bf g}_j^{(i)}$ means the corresponding value after passing the ReLU neuron. ${\bf l}_j^{(i)}$ and ${\bf u}_j^{(i)}$ refers to the lower bound and upper bound of the pre-activation ${\bf u}_j^{(i)}$ w.r.t. certain input perturbation. Fig. \ref{fig_unstable}(a) and Fig. \ref{fig_unstable}(d) are unstable neurons where ${\bf l}_j^{(i)}$ and ${\bf u}_j^{(i)}$ has different signs, while Fig. \ref{fig_unstable}(b) and Fig. \ref{fig_unstable}(c) are stable neurons where ${\bf l}_j^{(i)}$ and ${\bf u}_j^{(i)}$ has the same sign. The yellow areas in Fig. \ref{fig_unstable}(a) and Fig. \ref{fig_unstable}(d) refer to the bounded area of "triangle" relaxation and linear relaxation, respectively.

\subsection{What factors influence certified robustness?}

\label{sec_problem} Up to now, with state-of-the-art robust training method and certified verifier, the measurable bound of a trained non-linear neural network is influenced by two major factors:
\subsubsection{ The ground-truth bound of the network}

This is mainly decided by the training method. For example, certified training usually provides much higher certified robustness than adversarial training, which is demonstrated in our experiments. In this paper, we ideally hope pruning has positive or no influences on the normal training process, e.g. the influences of pruning-related regularizer to normal gradient back-propagation. 
Besides, the model size, i.e. the parameter number also matters, which is important in this paper since we can use pruning to reduce parameter number. Intuitively, with more parameter numbers, the bound of the network output tends to be looser. With network pruning, we can heuristically tighten the bound of the network since the network has fewer parameters. %\textcolor{blue}{I'm highly confused. ``training method" cannot alter model size because IMHO the latter is decided by arch. Pruning is also not a training method. I'm skeptical - is ``essential bound" or ``model capacity" the right term here? did you really inherit them from some paper (I'm never aware or) or coin them randomly?}

\subsubsection{Estimated bound tightness of the verifier}

This refers to the closeness of the estimated bound of the verification to the ground truth bound of the network, and it reflects the performance of the verifier. Since existing certified verifiers usually have a very large computational overhead to verify even a single sample, in practice, we are concerned about the bound tightness a verifier can reach given a limited verification time.

Although efficient certified verification for non-linear neural networks has made rapid progress in recent years, the bound tightness, running speed and affordable network capacity on limited hardware resources still have huge space awaiting to be improved. Specifically, we identify two major problems that influence the bound tightness:
\begin{itemize}
    \item {\bf Neuron Stability and Verification Speed.} As mentioned in Section 1, BaB is the main-stream framework of existing certified verification methods. The bound tightness is largely influenced by how many unstable neurons have been branched given a limited time. (Please refer to Section \ref{unstable} for the explanation of neuron stability).
    The verifier would need to visit more sub-domains by branching on unstable neurons within limited time, and thus the verification speed matters.
    \item{\bf Linear Relaxation.} To facilitate bound computation, many recent verifiers \cite{wang2021beta, xu2020fast, wong2018provable, zhang2018efficient, singh2019abstract},  utilize the linear relaxation method for unstable ReLU neurons, as we illustrated in Section \ref{unstable}. Normally, if an unstable neuron has not been branched by BaB, then this neuron gets linearly relaxed during bound propagation, which will also loosen the bound tightness of the verifier.
\end{itemize}

\subsection{What benefits for certified robustness can we expect from pruning?}
\label{benefits}

Network pruning can bring many benefits to problems as mentioned in Section \ref{sec_problem}, outlined as follows:
\begin{itemize}
    \item {\bf Reducing parameter number}. It can tighten bound propagation directly. Both structured and unstructured pruning can reduce parameter number and the bound propagation becomes tighter after pruning. Take the widely used bound propagation method for linear layers(e.g. convolution and full-connected layer)——Interval Bound Propagation (IBP) as an example, its computation can be formulated as follows:
\begin{equation}
\begin{split}
    \mu_{i-1}&=\frac{\bar{z}_{i-1}+\underline{z}_{i-1}}{2}\\
    r_{k-1}&=\frac{\bar{z}_{i-1}-\underline{z}_{i-1}}{2}\\
    \mu_k&={\bf W}\mu_{k-1}+b\\
    r_k&=|{\bf W}|r_{k-1}\\
    \underline{z}_{i}&=\mu_k-r_k\\
    \bar{z}_{i}&=\mu_k+r_k
    \end{split}
    \label{eq_ibp}
\end{equation}
    Eq. \ref{eq_ibp} computes the linear bound propagation for $i$th linear layer of the network, where $\bar{z}_{i-1}$ and $\underline{z}_{i-1}$ are the input lower and upper bounds of $i$th linear layer, and $\bar{z}_{i}$ and $\underline{z}_{i}$ the corresponding output lower and upper bounds, respectively. ${\bf W}$ and $b$ denote the weight and bias of the linear layer. The difference between output upper and lower bound equals $2|{\bf W}|r_{k-1}$. By network pruning, the weight matrix ${\bf W}$ becomes sparser, and thus the difference between output upper and lower bound tends to become smaller, and thus the overall output bound of the network would be tightened.
    \item {\bf Reducing unstable neurons}. By reducing the number of unstable neurons, we can reduce the number of linear relaxations and sub-domain splits needed by the verification process, which can also directly tighten the bound and accelerate the verification process as well. However, for most existing pruning methods, reducing unstable neurons is not an explicitly designated goal, but rather a possible side effect. %\textcolor{blue}{You were not clear this benefit is not essential nor natural by pruning. It is a side-product by most pruning, unless specially designed}
    \item{\bf Real hardware acceleration with structural sparsity}. If adopting structured pruning, we can eliminate channels which will concretely reduce the network width on the hardware implementation level. This can accelerate verification and even make resource-intensive verification possible on larger models. %\textcolor{blue}{Most reviewers will confuse this bullet and the first bullet. I'm against the current terms. IF you do plan to show the actual hardware-measured verification acceleration by channel pruning, MAKE this term clearly different from bullet 1, e.g. ``Real hardware acceleration benefits with structural sparsity".}
\end{itemize}
Those possible benefits are further entangled with each other. For empirical evaluation of these benefits, we simply follow the classical criteria: to evaluate the certified accuracy, time and memory consumption, and network width/depth that can be verified after pruning.

\subsection{Pruning methods}

In this section, we test a range of off-the-shelf pruning methods for improving certified robustness. For each pruning method, unless otherwise mentioned, we combine it with iterative pruning with weight rewinding \cite{frankle2018lottery}, as we find iterative pruning with weight rewinding generally enhances performance compared to finetune-based pruning or one-shot pruning in our experiments. 

\subsubsection{Existing pruning methods}

In unstructured pruning, for simplicity, we only prune the weights of convolutional layers and ignore linear layers. We pick several representative methods including: 1) {\bf Random Pruning}: pruning weights randomly, which is used for sanity check in our experiments. 2) Lottery ticket hypothesis (LTH), or denoted as {\bf IMP} \cite{frankle2018lottery}: pruning weights with the smallest magnitudes, the most standard pruning scheme.  3) {\bf SNIP} \cite{lee2018snip}: pruning weights with least loss sensitivity w.r.t. percentile magnitude change. 4) {\bf TaylorPruning (TP)} \cite{evci2018detecting}: saliency-based pruning via a first-order Taylor approximation. 5) {\bf HYDRA} \cite{sehwag2020hydra}: learnable mask-based pruning that minimizes the robustness loss empirically. 

For structured pruning, we choose two methods: 1) {\bf StructLTH} \cite{anonymous2022lottery}: a structured variant of LTH recently proposed, by using IMP first then ranking channels by their total magnitudes in remaining weights from high-to-low. Then we prune lowest-ranked channels and refill the IMP-pruned weights in the remaining channels. 2) {\bf Network Slimming} \cite{liu2017learning}: for batch normalization (BN) \cite{ioffe2015batch} layers, we have:
\begin{equation}
    y=\frac{x-E[x]}{\sqrt{Var[x]+\epsilon}}*\gamma+\beta
\end{equation}
Network slimming enforces the L1-norm regularizer on $\gamma$ and prune channels with the smallest $\gamma$ magnitudes.

\subsubsection{Stability-based Pruning}

 In the context of certified robustness, we wish to eliminate unstable neurons as much as possible, such that the estimated bound tightness of the verification can be improved. Pruning is a natural choice to accomplish this goal. Next, we first introduce a criterion that measures the degree of neuron stability of a given network, and then introduce an effective stability-based regularizer and corresponding unstructured and structured pruning methods.
 
Formally, we denote $[{\bf l}^{(i)}_j, {\bf u}^{(i)}_j]$ as {\it bound interval} of the pre-activation value of $j$-th ReLU neuron at $i$-th layer. We want to measure not only the number of unstable ReLU neurons with bound intervals crossing the zero point but also to what extent the instability is. To address this problem, we propose to use $-{\bf l}^{(i)}_j\cdot {\bf u}^{(i)}_j$ to measure the degree of instability of this neuron. It is easy to know that if we keep the bound interval width ${\bf u}^{(i)}_j-{\bf l}^{(i)}_j$ unchanged, then $-{\bf l}^{(i)}_j\cdot {\bf u}^{(i)}_j$ reaches maximum when ${\bf l}^{(i)}_j=-{\bf u}^{(i)}_j$, which means maximal instability given the same bound interval width. And thus, we simply use the average of this criterion to denote the total degree of instability of the network:

\begin{equation}
    instability = \sum\limits_{i}{\sum\limits_{j}{-{\bf l}^{(i)}_j\cdot {\bf u}^{(i)}_j}}
\end{equation}

Similar to this criterion, \cite{xiao2018training} proposed a regularizer named {\it RS Loss} to regularize ReLU stability and improved certified robustness. The RS Loss is defined as:

\begin{equation}
l_j^{rs} = -\tanh(1+{\bf l}_j\cdot{\bf u}_j)
\end{equation}

This loss can be naturally used to optimize the instability criterion, as the $tanh$ wrapper provides smooth gradients. However, we empirically find that for deep networks, batch-normalization (BN) \cite{ioffe2015batch} layers, which are placed before ReLU layers, are necessary for the training convergence of pruned subnetworks. In this way, the performance of the RS Loss regularizer is insignificant, because it passes gradients to $\gamma$ and affects the training process, and the optimization space is relatively small due to the BN constraint. However, we find the pre-BN bounds (i.e. the input bounds of BN layers) to be very flexible. Thus, instead of regularizing the pre-activation bounds using RS Loss, we propose an alternative of \textit{Normalized RS Loss} (\textbf{NRSLoss}) to directly regularize the pre-BN bounds, normalized from pre-activation bounds ($\gamma$ is the BN weight of the corresponding channel):
\begin{equation}
    l_j^{nrs} = -\tanh(1+\frac{{\bf l}_j\cdot{\bf u}_j}{\gamma^2}).
\end{equation}
Since the magnitudes of ${\bf l}_j$ and ${\bf u}_j$ are scaled from pre-BN bounds by the factor of $\gamma$, NRSLoss essentially computes RS Loss on the pre-BN bounds ${\bf l}_j/\gamma$ and ${\bf u}_j/\gamma$. During training, the NRSLoss is combined with the original loss with an empirical coefficient. Note that we {\bf stop} the gradients back-propagated from NRSLoss to $\gamma$, to ensure stable training, especially for pruned subnetworks.  

The loss landscape of NRSLoss is shown in Figure \ref{nrsloss}. Note that the {\it stability} term is essentially  $0-instability$. From the loss landscape, we can interpret NRSLoss from another perspective: it penalizes neurons with high instability and low channel importance; when the instability increases, it takes larger channel importance to suppress NRSLoss.

We empirically find that combining NRSLoss as a training regularizer with pruning weights based on the least weight magnitude criterion is most effective. We call this pruning scheme IMP+NRSLoss, which we interpret as follows: the model is trained based on a weighted combination of robustness loss and NRSLoss, and the magnitude of each trained individual weight reflects its saliency w.r.t. both robustness loss and NRSLoss, i.e. saliency w.r.t. robustness and stability, and by pruning weights with minimal magnitude-based saliency of robustness and stability, we can minimize the negative effects on robustness and stability brought by pruning, and benefit from positive effects of pruning to certified robustness as introduced above.

% For structured pruning, we prune channels with highest average NRSLoss:
%     \begin{equation}
%         l_{channel}^{nrs} = \frac{1}{N}\frac{1}{|channel|}\sum\limits_{N}{\sum\limits_{j\in channel}{l_j^{nrs}}}
%     \end{equation}
% where $|channel|$ is number of neurons of each channel, and $N$ is the number of samples in test set we use to get the statistical average of normalized RS loss. High value of $l_{channel}^{nrs}$ means the channel has high instability and small channel saliency (i.e. small $\gamma$).

We stress the two-fold novelty of NRSLoss as follows:
\begin{itemize}
    \item It takes into account both neuron importance and stability (see Figure \ref{nrsloss}). In NRSLoss, $\gamma$ is the corresponding channel weight of the Batch-Normalization layer, whose magnitude denotes the importance of each channel. The NRSLoss is high when the neuron is unstable and the corresponding channel has low importance. In contrast, the RSLoss is irrelevant to channel importance, which might lead to imposing too much regularization on important neurons.
    \item It also disentangles the influence of stability regularization with the BN layers, via eliminating the magnitude scaling effect to the pre-activation bounds brought by the channel weight $\gamma$. In this way, BN layers can still be learned normally to control the gradients.
%    gradient vanishing/exploding problem during training, which is important for the performance of the trained models.
\end{itemize}

\section{Experiments}

In this section, we evaluate all introduced pruning methods with different training schemes and perturbation scales, and try to address three major questions: \textit{(1) Can existing pruning methods improve certified robustness generally? (2) How can NRSLoss-based pruning improve certified robustness? (3) Can we find certified lottery tickets, i.e., sparse subnetworks after pruning that can restore not only the original performance but also certified robustness?} 

Based on our experiment findings, we also provide more ablation studies to further rationalize our claims. Finally, we briefly summarize our experimental findings with several interesting takeaways.

%In next subsections, we first introduce the experiment setups and then present the  experiment results and analysis.

%\textcolor{blue}{First, is CIFAR10 only ok? Second, after reading the entire Section 4, I feel lost about what the take-home message is. The paper's theme is to answer the title question; so now does sparsity help? does it help all robustness metrics? always or sometimes? how do I choose between different pruning methods list above? And most importantly, what is YOUR contribution and novelty as experiments show? You can take a look at the structure of: http://proceedings.mlr.press/v119/you20a/you20a.pdf}

\subsection{Experiment Setup}
\label{setup}

\begin{table*}[h]
\centering
\caption{Comparisons of subnetwork robustness and verification time for different training and pruning methods on FashionMNIST, SVHN, and CIFAR10 datasets. $std$, $adv$, $ver$, $t$ refer to standard accuracy(\%), adversarial accuracy(\%), verified accuracy(\%) and time consumption (s/sample), respectively. Note that the remain ratio refers to remaining weights for unstructured pruning, and remaining channels for structured pruning, respectively.}
\label{tb_2.255}
\begin{center}
\begin{small}
\begin{sc}
\begin{tabular}{c|c|c|c c c c |c|c c  c }
\toprule
&\multicolumn{2}{|c|}{Training Method}&\multicolumn{4}{c|}{FGSM} &  &\multicolumn{3}{c}{auto-LiRPA} \\
\hline
\makecell{Dataset}& \makecell{Pruning\\ Method}& \makecell{Remain\\ Ratio}
&$std$&$adv$  &$ver$ &$t$  
& \makecell{Remain\\ Ratio}
&$std$  &$ver$ &$t$ 
\\
\hline
\multirowcell{10}{FashionMNIST \\ $\epsilon=0.1$}& Dense&1 &{\bf 85.2} &81.2 &1.5 &298.3  &1 &77.2  &68.8 &7.23 \\
\cline{2-11}

%\multirowcell{7}{Unstru-\\ ctured}& Random& 0.03&51.2 &45.3 &{\bf 32.5} &39.82 &15.5&0.4 &55.5  & 44.0 &5.25 & 15.5    \\
& IMP &0.03 &80.2 &75.3 &39.0(\textcolor{green}{+37.5}) &85.9 &0.07 &78.1(\textcolor{green}{+0.9})  & 73.5(\textcolor{green}{+4.7}) &7.17    \\
& SNIP&0.03 &81.2 &77.3 &36.5(\textcolor{green}{+35.0}) &96.3  &0.80 &77.5(\textcolor{green}{+0.3}) &71.3(\textcolor{green}{+2.5}) &4.82     \\
& TP& 0.03&80.3 &76.1 &35.3(\textcolor{green}{+33.8}) &92.3 & 0.32 &79.4(\textcolor{green}{+2.2}) &73.5(\textcolor{green}{+4.7}) &6.60     \\
& HYDRA&0.03 &81.5 &77.2 &33.5(\textcolor{green}{+32.0}) &105.9  &0.51 &79.1(\textcolor{green}{+1.9}) &73.3(\textcolor{green}{+4.5}) &6.60    \\
& HYDRA+NRSLoss&0.03 &80.3 &76.7 & 36.5(\textcolor{green}{+39.5}) &105.3 &0.03 &81.5(\textcolor{green}{+4.3}) &74.2(\textcolor{green}{+5.4}) &8.90   \\
& IMP+RSLoss&0.03 &79.3 &72.2 &37.0(\textcolor{green}{+35.5}) &94.1  & 0.41 &{\bf 80.5}(\textcolor{green}{+3.3})  & 72.5(\textcolor{green}{+3.7}) &6.21     \\
& IMP+NRSLoss&0.03 &81.2 &74.9 &{\bf 41.5}(\textcolor{green}{+40.0}) &78.4  &0.05 & {\bf 80.5}(\textcolor{green}{+3.3})  &{\bf 74.0}(\textcolor{green}{+5.2}) & 6.06     \\
\cline{2-11}
& StructLTH[1]&0.35 & 80.5 & 78.3 &22.5(\textcolor{green}{+21.0}) &143.0&0.80 &80.0(\textcolor{green}{+2.8})  &72.4(\textcolor{green}{+3.6}) &6.14  \\
& Slim&0.35 &80.7 &78.3 & 31.0(\textcolor{green}{+29.5}) &105.0  &0.32 &78.5(\textcolor{green}{+1.3}) &71.8(\textcolor{green}{+3.0}) &2.96    \\
% & NRSLoss&0.35 &58.5 &54.5 &23.5 &93.5 &8.4 &0.79&57.0  &46.1 &4.27 &14.2    \\
\hline
\multirowcell{10}{SVHN \\ $\epsilon=2/255$}& Dense&1 &{\bf 94.8} &89.2 &2.0 &294.2  &1 & 76.3  &62.0 &6.68 \\
\cline{2-11}

%\multirowcell{7}{Unstru-\\ ctured}& Random& 0.03&51.2 &45.3 &{\bf 32.5} &39.82 &15.5&0.4 &55.5  & 44.0 &5.25 & 15.5    \\
& IMP &0.03 &92.3 &84.4 &31.0(\textcolor{green}{+29.0}) &188.3  &0.16 &82.0(\textcolor{green}{+5.7})  & 67.3(\textcolor{green}{+5.3}) &6.65    \\
& SNIP&0.03 &92.4 &84.3 &29.1(\textcolor{green}{+27.1}) &204.4  &0.32 &82.5(\textcolor{green}{+6.2}) &66.5(\textcolor{green}{+4.5}) &5.93     \\
& TP& 0.03&92.4 &84.2 &28.5(\textcolor{green}{+26.5}) &197.3 & 0.21 &83.0(\textcolor{green}{+6.7}) &67.3(\textcolor{green}{+5.3}) &5.12    \\
%& OBD& 0.03&60.4 &55.8 &23.5 &135.7 &15.5& 0.5 &55.0  &46.5 &4.78 &15.5    \\ %deleted since there are already several saliency based methods
& HYDRA&0.03 &91.2 &82.7 &{\bf 41.5}(\textcolor{green}{+39.5}) &146.3 &0.64 &79.5(\textcolor{green}{+3.2}) &66.3(\textcolor{green}{+4.3}) &8.75   \\
& HYDRA+NRSLoss&0.03 &89.8 &81.1 & 40.5(\textcolor{green}{+38.5}) &144.2 &0.55 &82.0(\textcolor{green}{+5.7}) &66.0(\textcolor{green}{+4.0}) &4.02   \\
& IMP+RSLoss & 0.03 & 85.6 & 75.9 & 25.2(\textcolor{green}{+23.2}) &179.2  & 0.03 &83.5(\textcolor{green}{+7.2})  & 65.7(\textcolor{green}{+3.7}) &6.52     \\
& IMP+NRSLoss& 0.03 & 92.1 &83.2 &33.5(\textcolor{green}{+31.5}) &153.2  &0.08 &{\bf 86.0}(\textcolor{green}{+9.7})  &{\bf 68.3}(\textcolor{green}{+6.3}) & 6.06    \\
\cline{2-11}
& StructLTH[1]&0.27&87.7 & 78.0 &33.0(\textcolor{green}{+31.0}) &159.1 &0.55 &82.2(\textcolor{green}{+5.9})  & 65.9(\textcolor{green}{+3.9}) &3.71  \\
& Slim&0.35 &89.7 &78.7 & 35.5(\textcolor{green}{+33.5}) & 151.8  & 0.63 & 84.0(\textcolor{green}{+7.7}) & 65.3(\textcolor{green}{+3.3}) &7.77    \\
% & NRSLoss&0.35 &58.5 &54.5 &23.5 &93.5 & 8.4 & 0.79 & 83.0  &46.1 &4.27 &14.2    \\

\hline
\multirowcell{10}{CIFAR10 \\ $\epsilon=2/255$}& Dense&1 &{\bf 82.4} &68.6 &1.5 &278.9 &1 &54.1  &43.0 &6.68   \\
\cline{2-11}

%\multirowcell{7}{Unstru-\\ ctured}& Random& 0.03&51.2 &45.3 &{\bf 32.5} &39.82 &15.5&0.4 &55.5  & 44.0 &5.25 & 15.5    \\
& IMP &0.03 &62.2 &55.4 &23.5(\textcolor{green}{+22.0}) &135.3 &0.13 &61.0(\textcolor{green}{+6.9})  & 50.1(\textcolor{green}{+7.1}) &6.65    \\
& SNIP&0.03 &61.5 &55.1 &22.5(\textcolor{green}{+21.0}) &128.4 &0.04 &59.8(\textcolor{green}{+5.7}) &48.4(\textcolor{green}{+5.4}) &6.67    \\
& TP& 0.03&59.7 &55.4 &24.0(\textcolor{green}{+22.5}) &132.4 & 0.05 &59.9(\textcolor{green}{+5.8}) &47.6(\textcolor{green}{+4.6}) &6.02     \\
%& OBD& 0.03&60.4 &55.8 &23.5 &135.7 &15.5& 0.5 &55.0  &46.5 &4.78 &15.5    \\ %deleted since there are already several saliency based methods
& HYDRA&0.03 &60.4 &55.4 &23.5(\textcolor{green}{+22.0}) &132.2  &0.11 &60.5(\textcolor{green}{+6.4}) &48.3(\textcolor{green}{+5.3}) &8.75   \\
& HYDRA+NRSLoss&0.03 &54.9 &48.4 &25.0(\textcolor{green}{+22.0}) &132.2  &0.05 &58.0(\textcolor{green}{+3.9}) &49.0(\textcolor{green}{+6.0}) &8.75   \\
& IMP+RSLoss&0.03 &60.2 &54.2 &23.5(\textcolor{green}{+22.0}) &134.4 & 0.13 &58.6(\textcolor{green}{+4.5})  & 46.3(\textcolor{green}{+3.3}) &6.52    \\
& IMP+NRSLoss&0.03 &60.7 &51.0 &25.0(\textcolor{green}{+23.0}) &131.2 &0.21 &{\bf 62.2}(\textcolor{green}{+8.1})  &{\bf 51.2}(\textcolor{green}{+8.2}) &6.06    \\
\cline{2-11}
& StructLTH[1]&0.35&55.6 & 48.3 &14.0(\textcolor{green}{+12.5}) &143.7 &0.55 &57.5(\textcolor{green}{+3.4})  &44.6(\textcolor{green}{+1.6}) &3.71 \\
& Slim&0.35 &56.9 &49.6 &{\bf 26.0}(\textcolor{green}{+24.5}) &72.9  &0.79 &59.2(\textcolor{green}{+5.1}) &47.5(\textcolor{green}{+4.5}) &5.65   \\
% & NRSLoss&0.35 &58.5 &54.5 &23.5 &93.5 &8.4 &0.79&57.0  &46.1 &4.27 &14.2    \\
\bottomrule 
\end{tabular}
\end{sc}
\end{small}
\end{center}
\end{table*}

\begin{table*}[h]
\centering
%\vspace{-0.5em}
\caption{Verified accuracies of different pruning and robust training methods and perturbation scales $\epsilon$ under auto-LiRPA setting.}
\label{tb_8.255}
\begin{center}
\begin{small}
\begin{sc}
\begin{tabular}{c|c|c|c c c| c| c c  c }
\toprule
\multicolumn{3}{c|}{$\epsilon$}&\multicolumn{3}{c|}{$2/255$}&&\multicolumn{3}{c}{$8/255$} \\
\hline
\makecell{Pruning\\ type}& \makecell{Pruning\\ Method}& \makecell{Remain\\ Ratio}
&$std$ &$ver$ &$t$&\makecell{Remain\\ Ratio}
&$std$ &$ver$ &$t$
\\
\hline
\multicolumn{2}{c|}{Dense}&1 &54.1 &43.0&6.68  & 1&36.1  &29.3 &5.83   \\
\hline
%\multirowcell{5}{Unstru-\\ ctured}& Random& 0.4& 55.0&44.0& 5.25&0.26 &37.5  &31.0 &2.74   \\
& IMP&0.13 &61.0 &50.1 &6.65  & 0.13 & 35.0   &29.8 &5.11    \\
& HYDRA&0.11 &60.4 &48.3 &8.75&0.11  &34.5  & 29.5 &6.36    \\
& IMP+RSLoss&0.13 &58.6 &46.3 &6.52 & 0.64 & 35.0 &28.0 &4.39    \\
& IMP+NRSLoss& 0.21 & {\bf 62.2} & {\bf 51.2} & 6.06 & 0.51 & {\bf 39.0}  &{\bf 31.5} &4.32  \\

\hline
\multirowcell{1}{Structured}& Slim&0.79  &59.2& 47.5  &5.65 &0.7 & 35.9 &29.8  & 3.86 \\
\bottomrule
\end{tabular}
\end{sc}
\end{small}
%\vspace{-0.5em}
\end{center}
\end{table*}

% \begin{table*}[h]
% \centering
% \caption{Performance comparisons of different pruning and robust training methods under 8/255 perturbation.}
% \label{tb_8.255}
% \begin{center}
% \begin{small}
% \begin{sc}
% \begin{tabular}{c|c|c|p{0.3cm}<{\centering} p{0.3cm}<{\centering} p{0.3cm}<{\centering} p{0.3cm}<{\centering} |c| p{0.3cm}<{\centering} p{0.3cm}<{\centering} p{0.3cm}<{\centering} p{0.3cm}<{\centering} }
% \toprule
% \multicolumn{3}{c|}{Training Method}&\multicolumn{4}{c|}{FGSM}&&\multicolumn{4}{c}{auto-LiRPA} \\
% \hline
% \makecell{Pruning\\ type}& \makecell{Pruning\\ Method}& \makecell{Remain\\ Ratio}
% &$std$&$adv$ &$ver$ &$t$&\makecell{Remain\\ Ratio}
% &$std$&$adv$ &$ver$ &$t$
% \\
% \hline
% \multicolumn{2}{c|}{Dense}& & && & & 1&38.0 &- &31.5 &3.7   \\
% \hline
% \multirowcell{5}{Unstru-\\ ctured}& Random& & && & &0.26 &37.5  &- &31.0 &2.74   \\
% & IMP& & & & & & 0.17 & {\bf 41.0}  &- &32.0 &7.39    \\
% & HYDRA& & & &&&0.10  &38.5 &-  & 31.5 &6.27    \\
% & IMP+RSLoss& && & & & 0.64 & 37.0 &- &28.5 &3.60    \\
% & IMP+NRSLoss& & && & &0.50 &40.5 &- &{\bf 33.5} &3.60   \\

% \hline
% \multirowcell{2}{Struc-\\ tured}& Slim& & & && & & & &  \\
% & NRSLoss& & & & & && & &  \\
% \bottomrule
% \end{tabular}
% \end{sc}
% \end{small}
% \end{center}
% \end{table*}

\subsubsection{Dataset and Network architecture}

Across our experiments, we use FashionMNIST \cite{xiao2017fashion}, SVHN \cite{netzer2011reading}, and CIFAR10 \cite{krizhevsky2009learning} as the benchmark datasets. We introduce them as follows: 
\begin{itemize}
    \item {\bf FashionMNIST}: FashionMNIST is an MNIST-like greyscale image classification dataset by replacing hand-written digits with fashion items, which are more difficult to classify. It has a training set of 60,000 examples and a test set of 10,000 examples. Each example is a 28x28 grayscale image, associated with a label from 10 classes. We use the first 200 samples from the testing dataset for verification.
    \item {\bf SVHN}: SVHN is a dataset consisting of Street View House Number images, with each image consisting of a single cropped digit labeled from 0 to 9.  Each example is a 32x32 RGB image. We use the first 200 samples from the testing dataset for verification.
    \item {\bf CIFAR10}: CIFAR10 is a dataset consisting of 10 object classes in the wild, and each class has 6000 samples. This dataset is commonly used in prior works in complete verification, following \cite{wang2021beta}, we choose the ERAN test set \cite{singh2019abstract} which consists of 1000 images from the CIFAR10 test set. Note that we only use the first 200 samples in the ERAN test set for verification efficiency.
\end{itemize}

\begin{figure}[!h]
    \centering
    %\vspace{-0.3em}
    \includegraphics[width=\linewidth]{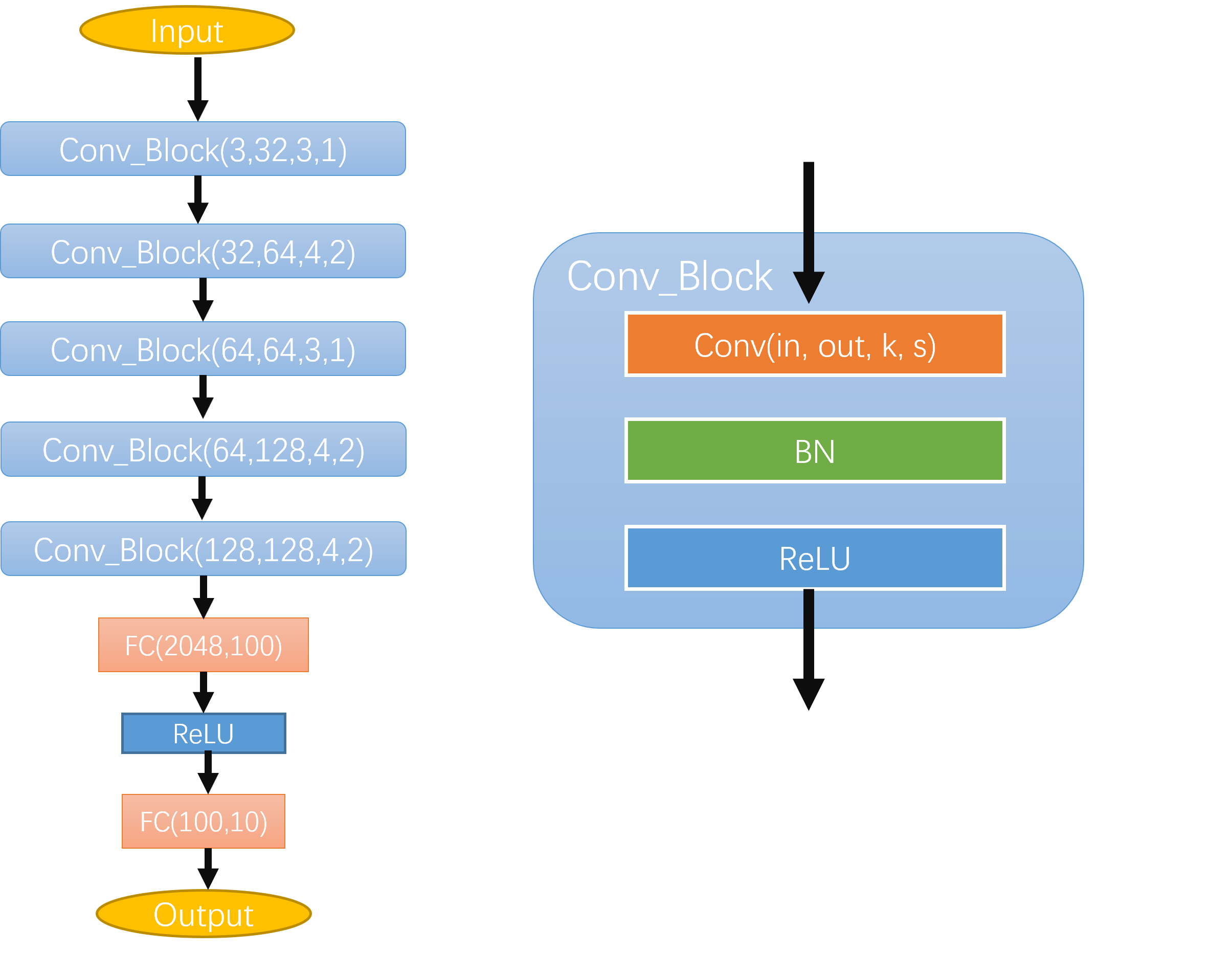}
     %\vspace{-0.3em}
    \caption{The illustration of the feedforward model architecture in our experiments.}
     %\vspace{-0.3em}
    \label{fig_network}
\end{figure}

We use a 7-layer feed-forward convolutional neural network as the benchmark model, whose architecture is shown in Table \ref{tb_arch} and Figure \ref{fig_network}. The design follows the {\it cifar10-model-deep} setting in \cite{wang2021beta}, but is wider, deeper, and has BN layers. This model is the largest network that can be fitted in a GPU with 24GB memory for complete verification.

\subsubsection{Pruning methods}

We try different setups of hyperparameters for unstructured and structured pruning. For unstructured pruning methods, we follow the default setting in \cite{frankle2018lottery} and set the iterative weight pruning rate to 0.2, and prune 16 times with re-training; for structured pruning methods, we keep a similar pruning speed. \footnote{Note that for HYDRA pruning, the semi-supervised training scheme which exploits extra unlabeled data as in \cite{sehwag2020hydra} is NOT used in our experiments, for a fair comparison.}

We set the NRSLoss weight to 0.01 and set the L1-norm regularizer weight to 0.0001 following \cite{liu2017learning}.
For RS Loss and NRS Loss-based unstructured pruning, we train with these losses and pruning with IMP, as Section 3 introduced, denoted as {\it IMP+RSLoss} and {\it IMP+NRSLoss}. We also test training with NRSLoss and pruning with HYDRA, denoted as HYDRA+NRSLoss, to demonstrate the effectiveness of NRSLoss as a regularizer.

\begin{figure}[h]
  \centering
  \includegraphics[width=0.5\textwidth]{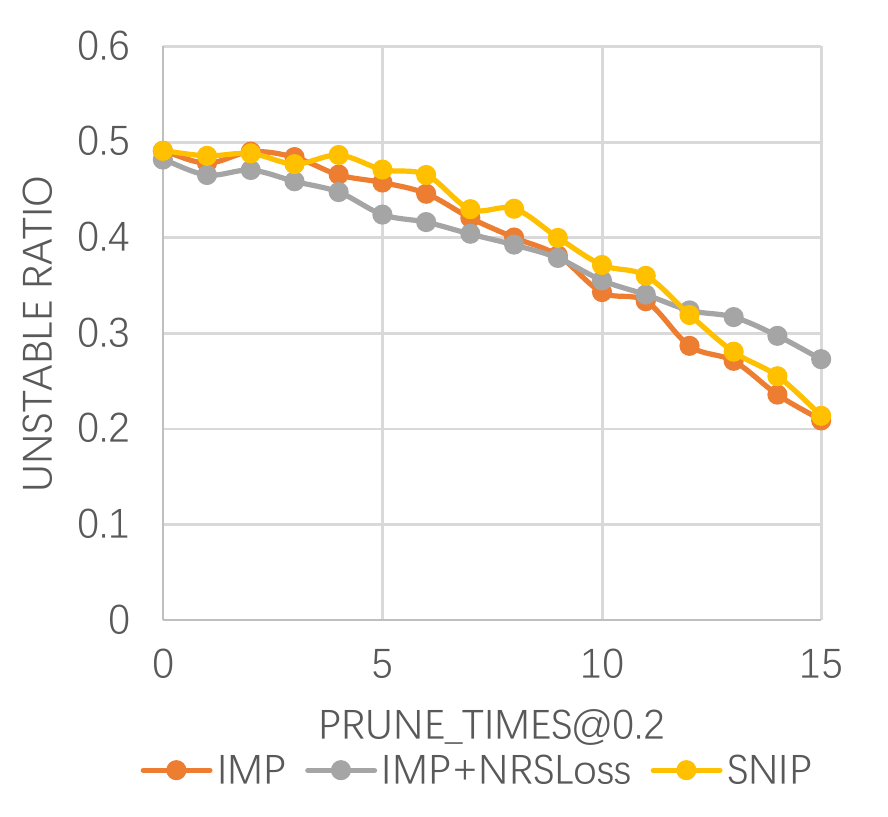}
  %\vspace{-0.3cm}
  \caption{The ratio of unstable neurons v.s. pruning times of different pruning methods under FGSM setting.}
  %\vspace{-0.3cm}
  \label{unstable_ratio}
\end{figure}

\subsubsection{Training methods}
\label{training methods}
To demonstrate the general effectiveness of network pruning, we choose SOTA adversarial and certified training methods:

     {\bf Adversarial training}: we choose the advanced FGSM+GradAlign \cite{andriushchenko2020understanding} as the adversarial training method (we denote it as FGSM for conciseness hereinafter). The learning rate is set to 0.01, and we use Stochastic Gradient Descent with 0.9 momentum and 0.0005 weight decay as optimizer. All GradAlign-related hyperparameters follow  \cite{andriushchenko2020understanding}.
     
 {\bf Certified training}: we choose auto-LiRPA \cite{xu2020automatic} under CROWN-IBP + Loss Fusion setting. The learning rate is set to 0.001, and we use Adam with a weight decay of 0 for RSLoss and NRSLoss-based pruning and 0.00001 for other pruning methods.
    For the bound computation of NRSLoss and RSLoss, we use the bound produced by auto-LiRPA during certified training and use IBP during adversarial training.  auto-LiRPA essentially uses IBP to compute bounds when input perturbation reaches a maximum during training and uses IBP constantly during testing. As we find the results to be unstable for certified training, we use five different random seeds to initialize the training process, and then average the results of the same iterations. This is different from adversarial training where we only run one experiment.% with seed 100.

FGSM and auto-LiRPA share some common hyperparameters. The batch size is set to 128, and we clip the norm of gradients to a maximum of 8. We train 200 epochs in one pruning iteration for each experiment, with a learning rate decay factor of 0.1 at 140 and 170 epochs. We set the input perturbation $\epsilon$ as $L_{inf}$-norm ball to 0.1 for the FashionMNIST dataset and 2/255 for SVHN and CIFAR10 datasets, and gradually increase $\epsilon$ from 0 to 2/255 starting from 11th epoch and until 80th epoch. We also scale the perturbation to 8/255 to validate the effectiveness of pruning under bigger perturbations. The training epochs under 8/255 perturbation is set to 300. After each pruning iteration, we rewind the remaining weights to initial states and reset the optimizer with the initial learning rate and $\epsilon$.

\begin{table}[h]
\centering
\caption{The feedforward model architecture in our experiments. ConvBlock($in$,$out$,$k$,$s$) refers to the composition of (convolution layer, BN layer, ReLU layer) where the convolution layer has $in$ input channels, $out$ output channels, $k\times k$ kernel size and $s$ strides. Note that for the FashionMNIST task which takes greyscale images instead of RGM images as input, we modify the input channel number of the first convolutional layer from 3 to 1, and modify the input units of the first FC layer from 2048 to 1152 accordingly.}
\label{tb_arch}
\begin{center}
\begin{small}
\begin{sc}
\begin{tabular}{c}
\toprule
Input \\
\hline
ConvBlock(3,32,3,1) \\
ConvBlock(32,64,4,2)   \\
ConvBlock(64,64,3,1)   \\
ConvBlock(64,128,4,2)    \\
ConvBlock(128,128,4,2)  \\
FC(2048,100)   \\
ReLU  \\
FC(100,10) \\
\hline
Output \\
\bottomrule
\end{tabular}
\end{sc}
\end{small}
\end{center} 
\end{table}

We replace the standard IMP with weight rewinding \cite{frankle2018lottery} with each training method and each pruning method, and output the pruned subnetworks with different sparsity during iterative pruning and re-training.

\subsubsection{Verifier and Evaluation Criterion}

For each training method and each pruning method, we use the SOTA certified verifier Beta-CROWN \cite{wang2021beta} to obtain the final accuracy of the subnetworks. Beta-CROWN is a highly GPU-parallelized verification framework and has SOTA performance in terms of bound tightness and verification speed. We choose ERAN benchmark \cite{bak2021second} which contains 1000 test images, and test on the first 200 images to reduce the time budget. We set the timeout of each test image to 300 seconds.
We test the standard, adversarial, and verified accuracies, as well as time and GPU memory consumption of each model. We run the verifications using one NVIDIA RTX A6000 GPU card.

\begin{figure*}[ht]
\centering
     \begin{subfigure}{.3\linewidth}
         \includegraphics[width=1.1\linewidth]{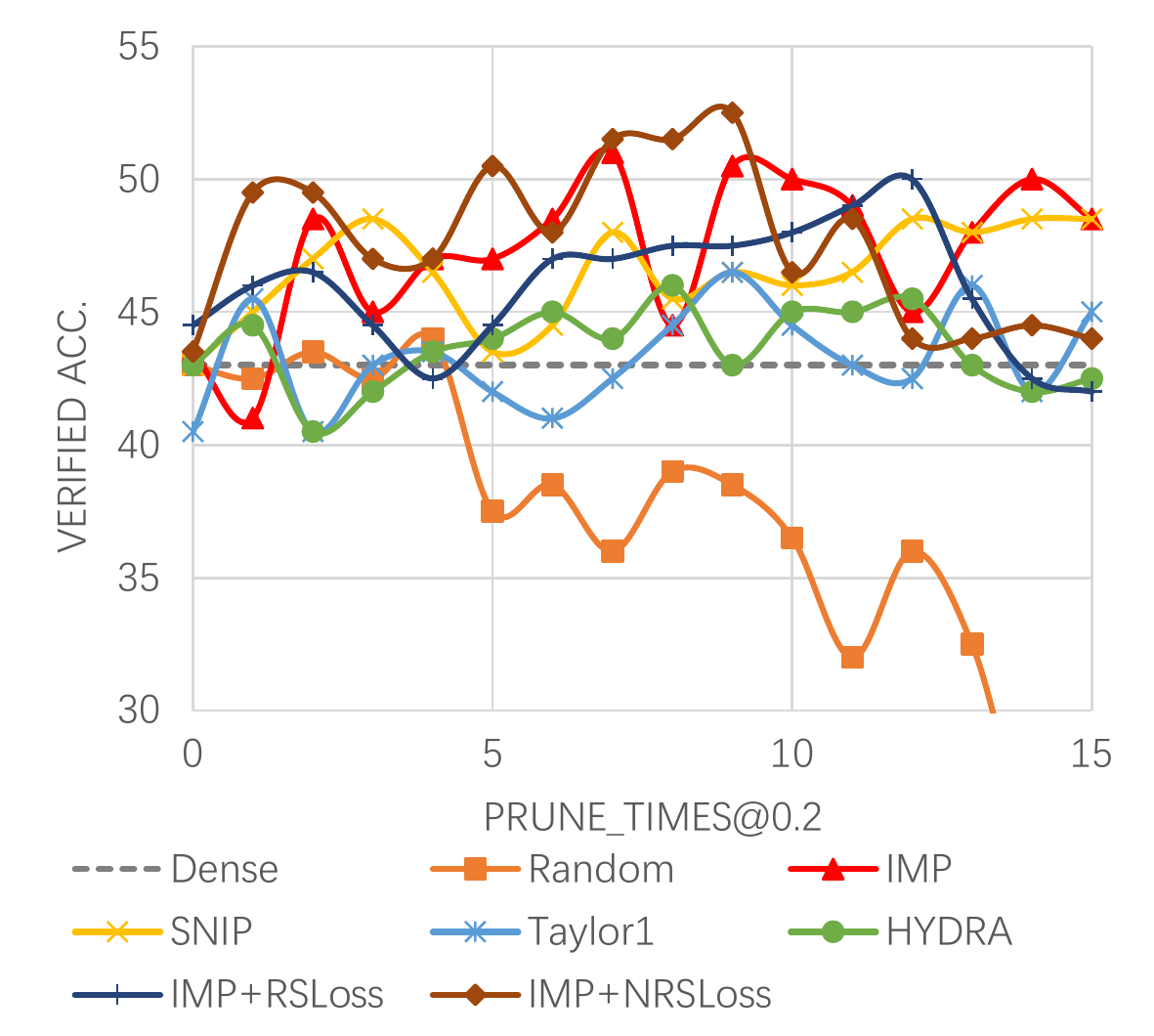}
         \caption{auto-LiRPA unstructured}
    \label{Time}
     \end{subfigure}
     \hfill
     \begin{subfigure}{.3\linewidth}
         \includegraphics[width=1.1\linewidth]{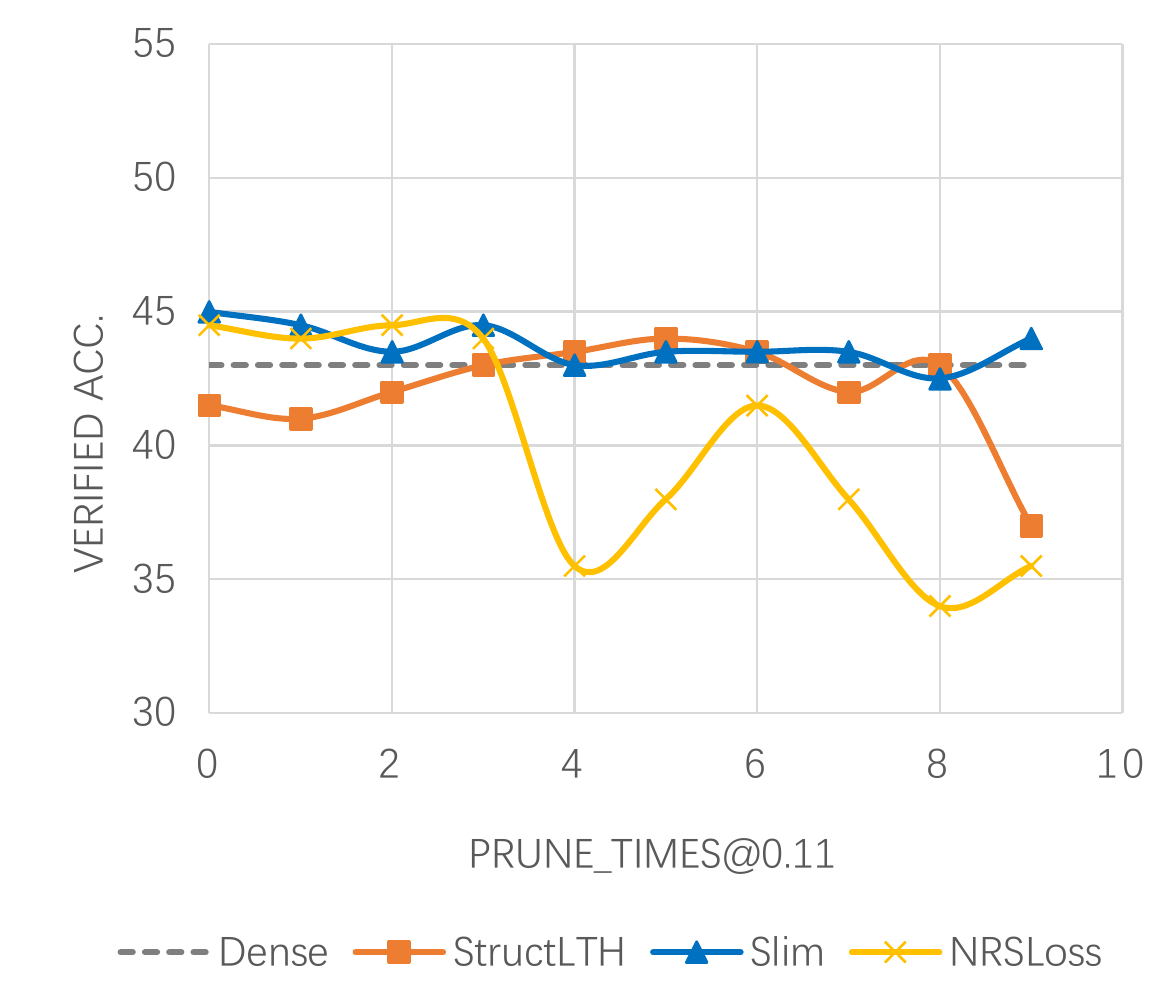}
         \caption{auto-LiRPA structured}
    \label{Memory}
     \end{subfigure}
     \hfill
     \begin{subfigure}{.3\linewidth}
         \includegraphics[width=1.1\linewidth]{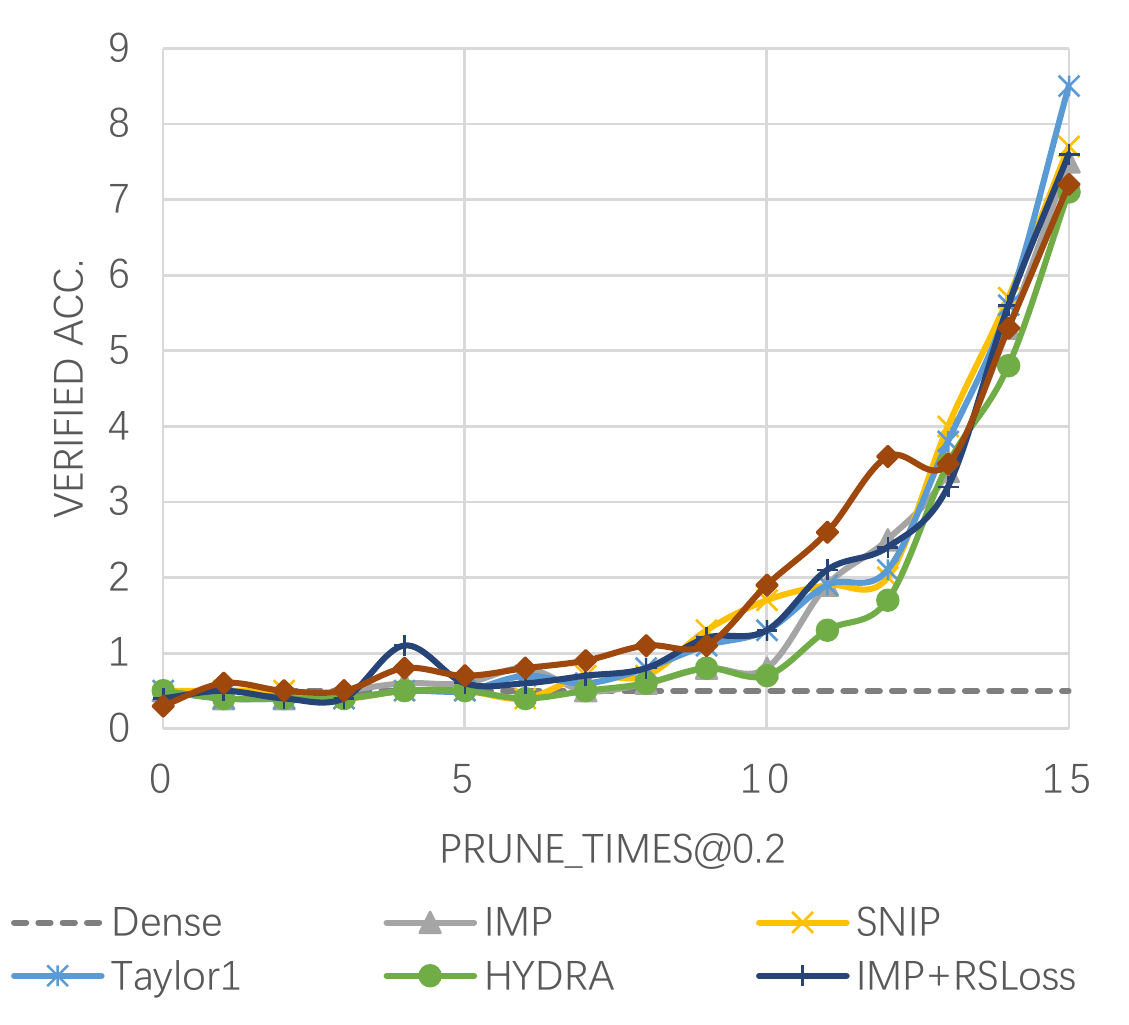}
         \caption{FGSM unstructured}
    \label{Memory}
     \end{subfigure}
        \caption{Verified Accuracy v.s. iterative pruning times on CIFAR10 dataset. (a) is unstructured pruning under auto-LiRPA training, (b) is structured pruning under auto-LiRPA training, (c) is unstructured pruning under FGSM. We omit structured pruning under FGSM due to page limit. Note that (c) is plotted using CROWN verifier instead of Beta-CROWN due to long verification time on Beta-CROWN.}
         %\vspace{-0.3em}
        \label{training curves}
\end{figure*}

\subsection{Experiment Results and Analysis}

% \begin{table}[h]
% \centering
% \label{tb_svhn}
% \caption{The standard and verified accuracies for models trained with auto-LiRPA and pruned with different pruning methods on SVHN dataset.}
% \begin{center}
% \begin{small}
% \begin{sc}
% \begin{tabular}{c|c|c|c}
% \toprule
%  & Sparsity & std acc. & ver acc. \\
% \hline
% $\epsilon$ & \multicolumn{3}{c}{$2/255$} \\
% \hline
% dense &1 &76.3 & 62.0   \\
% IMP & 0.16 &82.0 & 67.3  \\
% HYDRA &0.64 &79.5 & 66.3 \\
% HRank &0.21 & 80.5 & 66.0 \\
% IMP+RSLoss &0.03 & 83.5 & 65.7 \\
% IMP+NRSLoss & 0.08 &{\bf 86.0} & {\bf 68.3} \\
% SLIM & 0.63 & 84.0 & 65.0 \\
% \bottomrule
% \end{tabular}
% \end{sc}
% \end{small}
% \end{center}
% \end{table}

For the 3 benchmark datasets, the comprehensive experiment results under $\epsilon=2/255$ are shown in Table \ref{tb_2.255}, and the sample curves of Verified Accuracy v.s. iterative pruning times with seed 100 are shown in Figure \ref{training curves}. For each model, we verify all subnetworks produced by iterative pruning and report the best-verified accuracy and other corresponding evaluation metrics. We then choose several representative pruning methods that either have good performance or distinctive motivations or are important baselines, and evaluate them under different perturbation scales on the CIFAR10 dataset, as shown in Table \ref{tb_8.255}. For a fair comparison with HYDRA which is SOTA robustness-based pruning, we also reproduce a similar CROWN-IBP based experiment from \cite{sehwag2020hydra} as shown in Section \ref{hydra_lwm} to demonstrate that IMP can indeed outperform HYDRA. We also conduct a experiment to validate that weight-rewinding \cite{frankle2018lottery} is better than finetuning \cite{sehwag2020hydra} for improving certified robustness, as shown in Section \ref{finetune_rewind}. The results of Random Pruning are omitted from Table \ref{tb_2.255} and \ref{tb_8.255} since its standard accuracies are very poor and non-competitive (some can be found in Figure \ref{training curves} (a) for illustration purpose). We next present the result analysis.

\subsubsection{Can existing pruning methods improve certified robustness?}

From Table \ref{tb_2.255} and Table \ref{tb_8.255}, we observe general improvements in certified robustness brought by pruning, both under FGSM and auto-LiRPA settings. Specifically, Table \ref{tb_2.255} shows that under the auto-LiRPA setting, existing pruning methods can improve verified accuracies for $2.5-5.2\%$ on FashionMNIST, $3.3-6.3\%$ on SVHN, $1.6-7.1\%$ on CIFAR10, respectively and improve standard accuracies for $0.3-3.3\%$ on FashionMNIST, $3.2-9.7\%$ on SVHN, $3.4-8.1\%$ on CIFAR10 respectively, among which IMP consistently outperforms other existing pruning methods, with highest improvements of both standard and verified accuracies. This demonstrates that pruning can generally improve certified robustness. Moreover, under certified training, this improvement comes with no extra trade-off such as standard accuracy. We also observe that on more realistic datasets (SVHN, CIFAR10), the improvements under certified training are significantly bigger than that of the synthetic dataset (FashionMNIST).  

 Under the FGSM setting, there are great improvements in verified accuracies with different pruning methods ranging from $21.0-40.0\%$ on FashionMNIST, $23.2-39.5\%$ on SVHN, and $12.5-24.5\%$ on CIFAR10, respectively, among which IMP+NRSLoss, HYDRA and Network Slimming (Slim) obtain highest verified accuracy on FashionMNIST, SVHN, and CIFAR10, respectively. We also observe an obvious trade-off of standard/adversarial accuracy v.s. verified accuracy, i.e. with the big increase of verified accuracy after pruning, the standard and adversarial accuracies drop significantly. To explain this trade-off, we visualize the ratio of unstable neurons of different pruning methods, as shown in Figure \ref{unstable_ratio}. We find that the ratio of unstable neurons generally decreases as the sparsity gets higher, this is compliant with that neuron stability is important for certified robustness. However, if all neurons become stable, the whole network will become a linear function, which in turn withholds the standard accuracy. Hence the standard/verified accuracy trade-off is essentially the stability/expressiveness trade-off of the network. Nevertheless, this trade-off is not obvious under the auto-LiRPA setting since the training objective of auto-LiRPA incorporates standard accuracy. 
 
 Across different datasets, we observe general improvement brought by pruning for certified robustness, which consolidates our conclusion that pruning can generally improve certified robustness. In particular, NRSLoss-based pruning can outperform other pruning methods consistently under certified training and achieves competitive performance under adversarial training, which demonstrates the effectiveness of NRSLoss regularizer and the pruning scheme of IMP+NRSLoss, as we explained in the methodology.

%  By comparing FGSM and auto-LiRPA results, an interesting phenomenon emerges that random pruning has highest verified accuracy under adversarial training but trivial verified accuracy under certified training. This can be explained by viewing all selective pruning methods but random pruning as {\it partial-ordering} \cite{xu2011sparse}. Following the perspective of \cite{xu2011sparse}, network stability (i.e. certified robustness) originates from the ensemble from over-complete basis functions (here, subnetworks), and partial ordering (i.e. selective pruning) encourages the network to select distinct bases, hence the network stability will decrease. Since random pruning is not selective pruning, it can avoid this problem. Meanwhile in our experiments, basically all selective pruning methods but random pruning aim to preserve training objective. Under certified training, the training objective is certified robustness (i.e. network stability) which is also the objective of selective pruning, hence the performance of selective pruning is better than random pruning; however, under adversarial training, the training objective is adversarial accuracy, which is something different from certified robustness, and the disadvantage of selective pruning (i.e. partial-ordering) emerges, hence the random pruning which performs trivially in certified training can outperform selective pruning in adversarial training. %Overall speaking, every pruning method has improvement compared to dense models due to the reduction of parameter numbers. 

\begin{figure*}[!h]
\centering
     \begin{subfigure}{.4\linewidth}
         \includegraphics[width=1\linewidth]{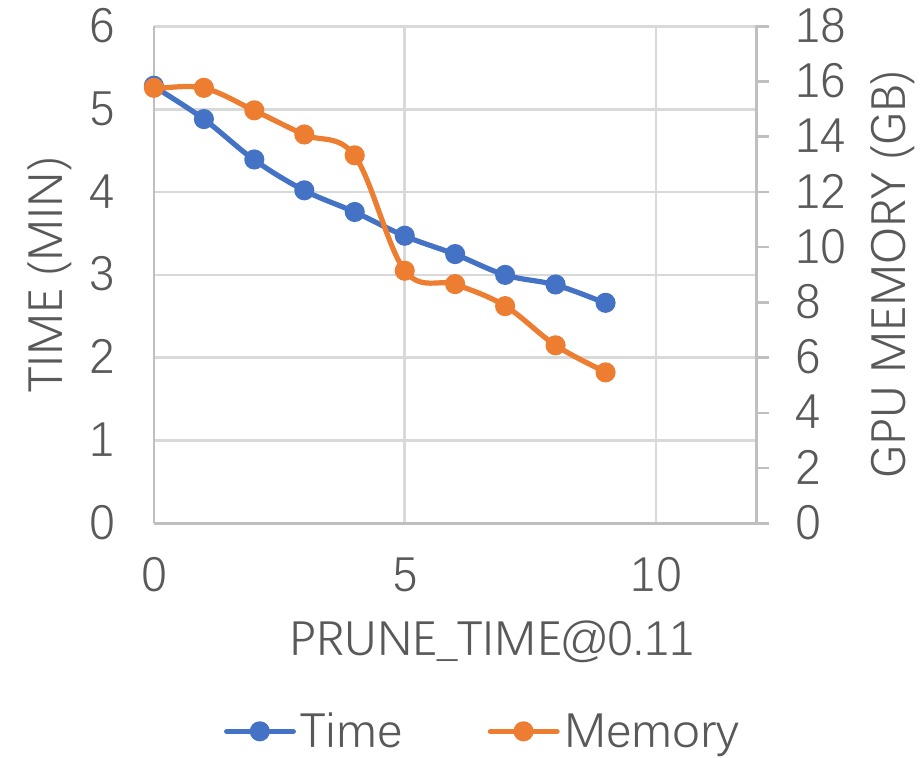}
         \caption{StructLTH w/ FGSM}
     \end{subfigure}
     \begin{subfigure}{.4\linewidth}
         \includegraphics[width=1\linewidth]{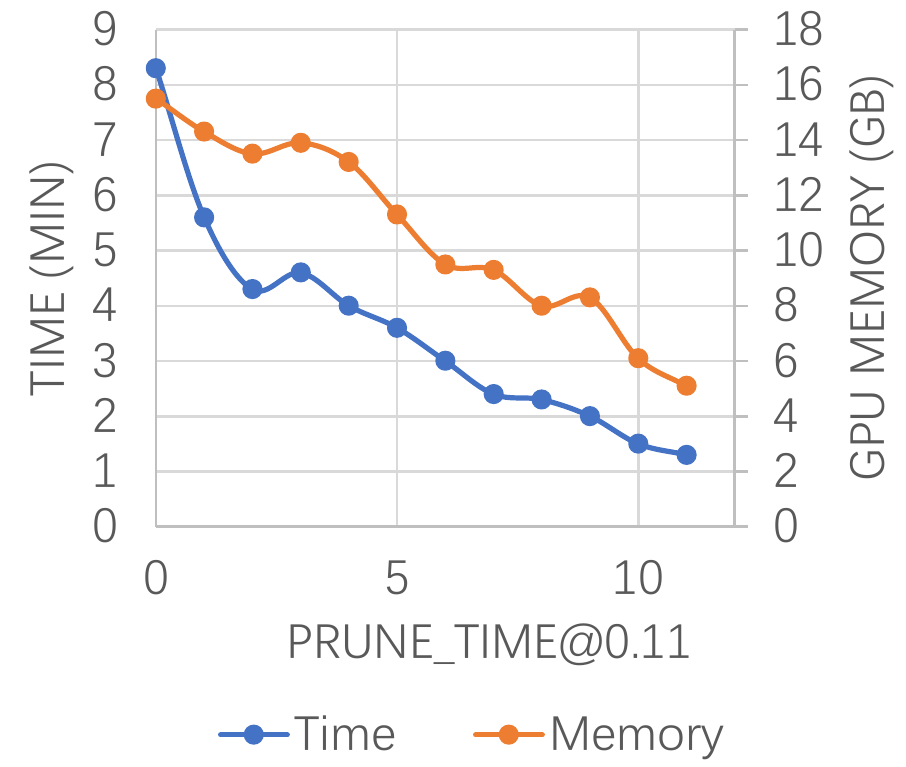}
         \caption{StructLTH w/ auto-LiRPA}
     \end{subfigure}
     
     \begin{subfigure}{.4\linewidth}
         \includegraphics[width=1\linewidth]{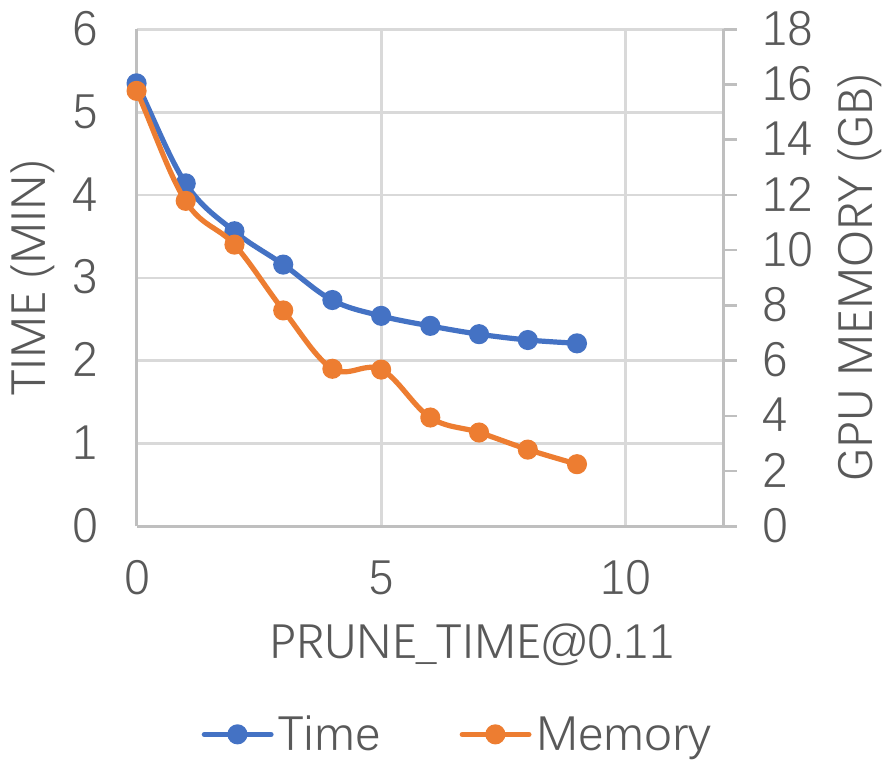}
         \caption{SLIM w/ FGSM}
     \end{subfigure}
     \begin{subfigure}{.45\linewidth}
         \includegraphics[width=1\linewidth]{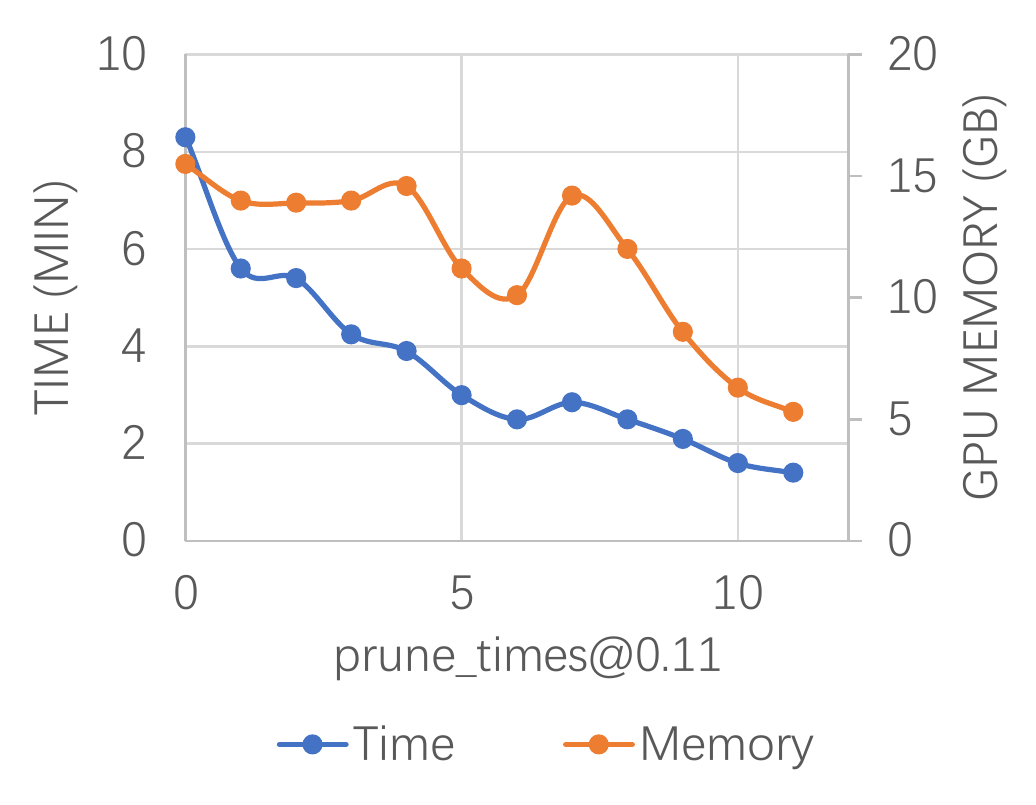}
         \caption{SLIM w/ auto-LiRPA}
     \end{subfigure}
        \caption{The mean verification time and average GPU memory consumption for structured pruning methods on CIFAR10 dataset under CROWN mode. X-axis means the number of pruning iterations with 0.11 channel pruning rate. Note that we don't do such test on complete verification of Beta-CROWN mode which is time consuming but has highly similar trend to the results under CROWN mode.}
        \label{fig_resource}
\end{figure*}

\begin{figure*}[!h]
\centering
     \begin{subfigure}{.4\linewidth}
         \includegraphics[width=1\linewidth]{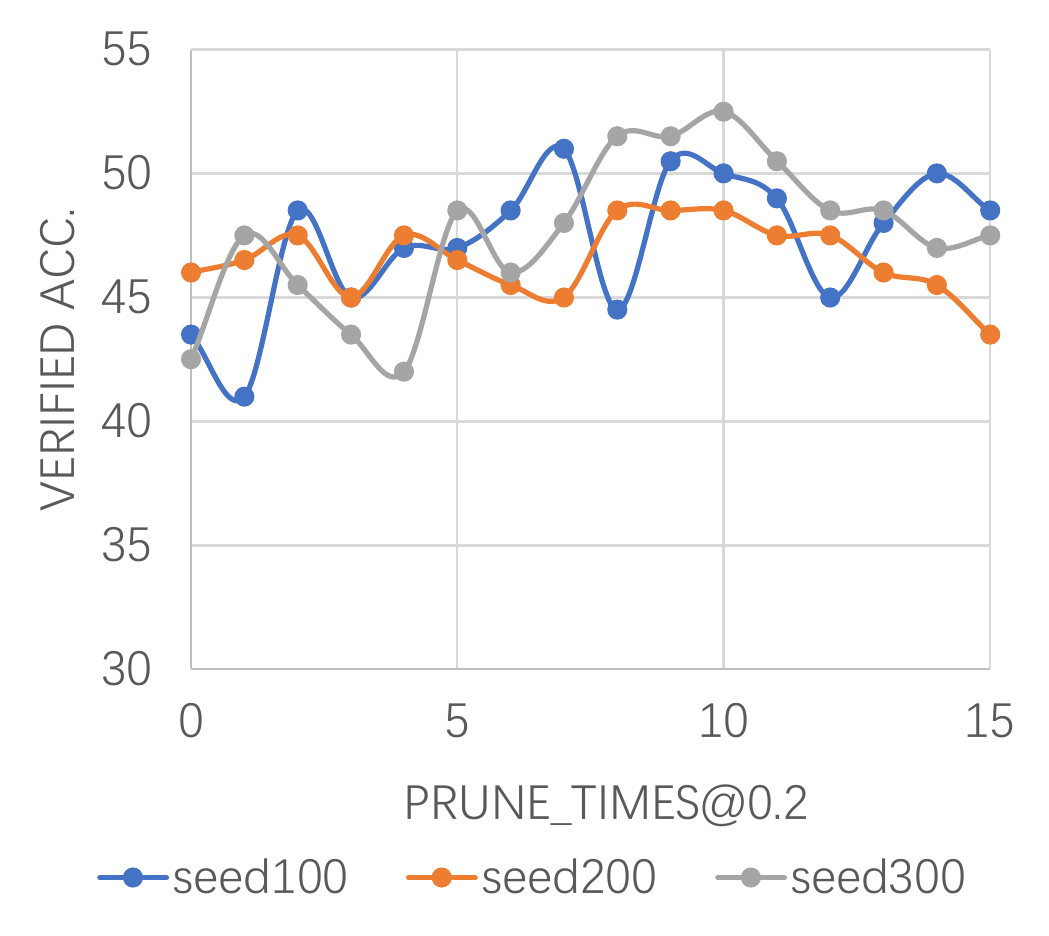}
         \caption{IMP}
     \end{subfigure}
     \begin{subfigure}{.4\linewidth}
         \includegraphics[width=1\linewidth]{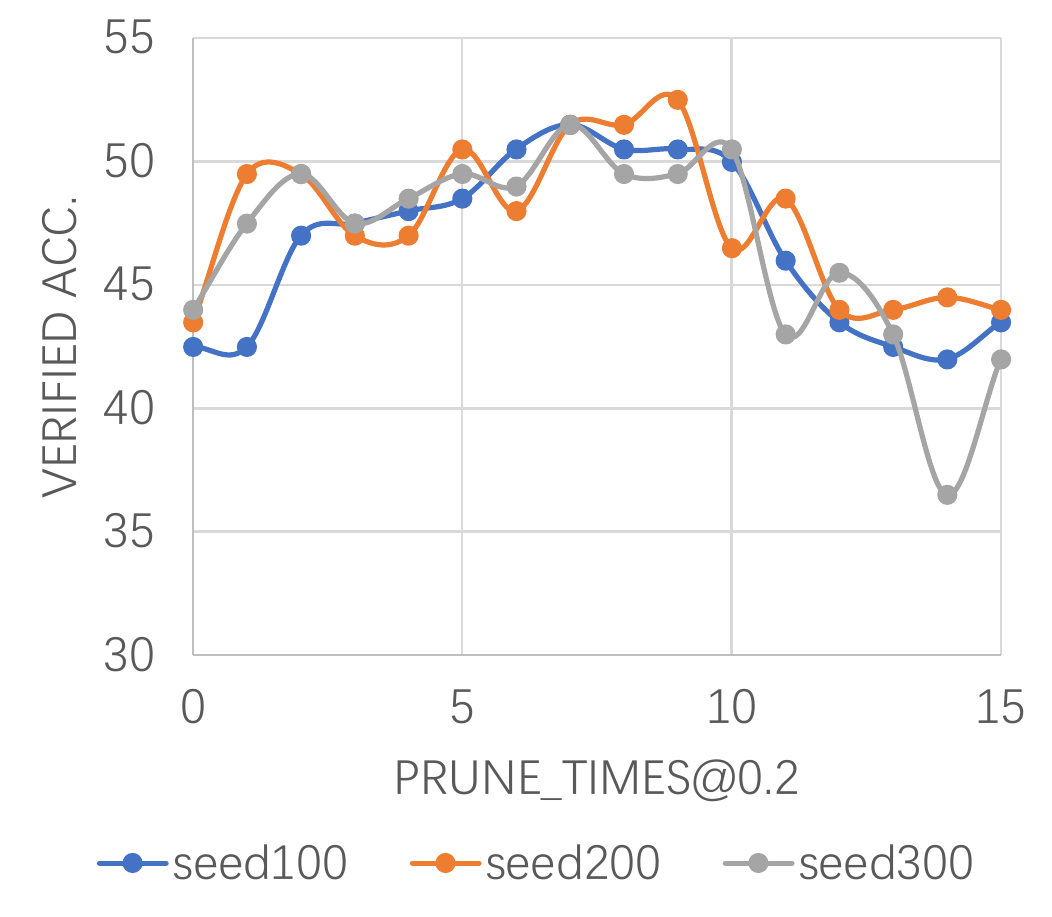}
         \caption{IMP+NRSLoss}
     \end{subfigure}
        \caption{Verified accuracy v.s. iterative pruning times under different random seeds with auto-LiRPA training. (a) is IMP pruning and (b) is IMP+NRSLoss pruning.}
        \label{seed curves}
\end{figure*}

{\bf Resource Consumption}: It can be observed from Table \ref{tb_2.255} and \ref{tb_8.255} that unstructured pruning tends to produce better performance than structured pruning. However, structured pruning has the advantage over unstructured pruning that it brings real hardware acceleration for certified verification, especially given that the computational overhead is a significant bottleneck for verifying large neural networks even with highly GPU-parallelized verifiers such as Beta-CROWN. We show an overview of the time and peak GPU memory consumption of structured pruning under different pruning stages as in Figure \ref{fig_resource} in Section. We can see that with every 3 pruning, which increases about 30\% channel sparsity, the time consumption for models trained with auto-LiRPA can reduce by about 50\%, whereas for models trained with FGSM can reduce by about $60-80\%$. We also observe that GPU memory consumption can be greatly reduced at high channel sparsity. The reduction in GPU memory consumption is even more important given the GPU memory bottleneck for complete verification of large neural networks. Furthermore, for these pruning methods, we observe similar high performing sparsity under different random seeds as shown in Figure \ref{seed curves} in Section, which means we do not need to verify every sparsity one by one to pick out the best sparsity, and it is crucial for accelerating the verification process in practice.

\subsubsection{How does NRSLoss-based pruning outperform other pruning methods?}
\label{how_does_nrsloss}

\begin{figure}[!h]
    \centering
    %\vspace{-0.3em}
    \includegraphics[width=\linewidth]{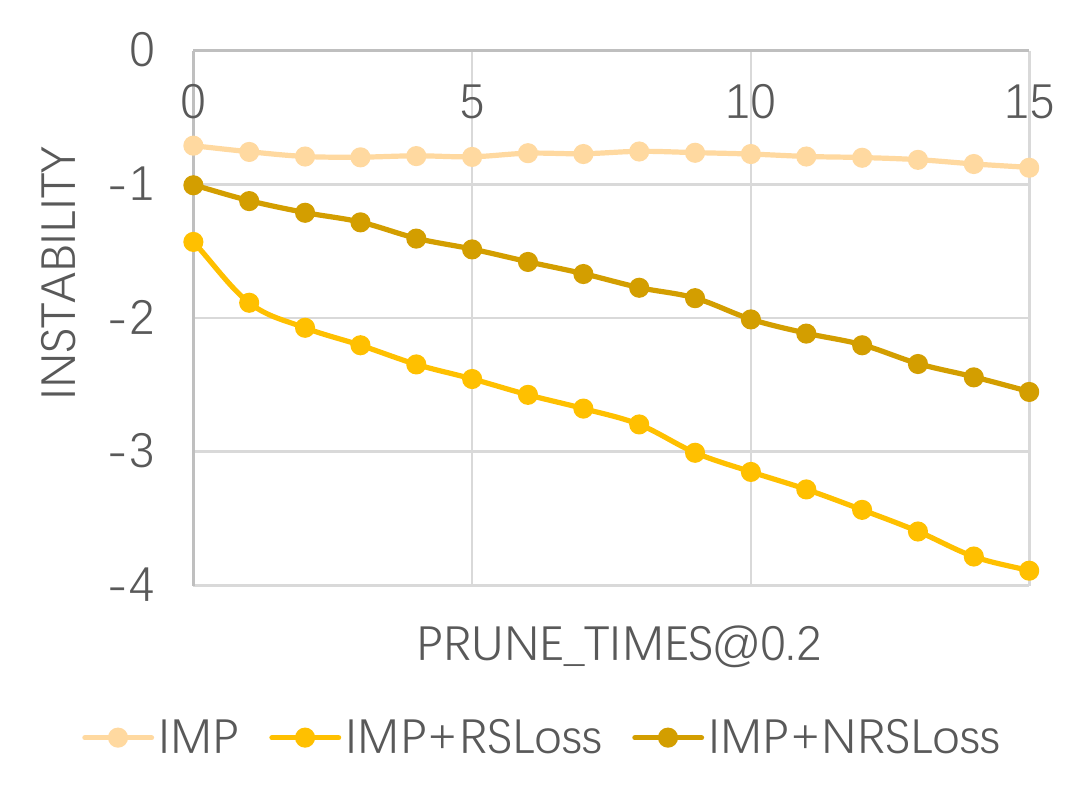}
     %\vspace{-0.3em}
    \caption{Network instability v.s. iterative pruning times of pre-activation and pre-BN forward pass values. The bounds are computed under 2/255 input perturbation using auto-LiRPA and the whole test set of CIFAR10.}
     %\vspace{-0.3em}
    \label{bound prod}
\end{figure}

From Table \ref{tb_2.255} and \ref{tb_8.255}, we observe that IMP+NRSLoss outperform other pruning methods under auto-LiRPA setting. Take the CIFAR10 dataset as an example, with 2/255 perturbation, IMP+NRSLoss improves certified accuracy for 8.2\% and standard accuracy for 8.1\%; with 8/255 perturbation, IMP+NRSLoss improves certified accuracy for 2.9\% and standard accuracy for 2.3\%. Notably, IMP+NRSLoss achieves both the highest standard and verified accuracies, since the training objective of auto-LiRPA incorporates standard accuracy. By comparing  HYDRA setting and HYDRA+NRSLoss setting, we observe NRSLoss can improve the verified accuracy for HYDRA pruning in FashionMNIST and CIFAR10 dataset, and has better standard/verified accuracy trade-off under certified training for SVHN dataset. We thus conclude that NRSLoss regularizer is effective for HYDRA pruning in most cases and is effective for IMP pruning for all cases we have tested. To demonstrate that the performance improvements of NRSLoss indeed come from stability-based regularization as discussed in Section 3, we visualize the pre-activation {\it network instability} (as proposed in Section 3) in Figure \ref{bound prod}. We observe that the RS Loss and NRS Loss-based pruning have significantly lower instability compared to IMP, and the instability decreases as the sparsity gets higher, which proves that pruning with NRSLoss and RSLoss regularizer can decrease network instability, hence improving the certified robustness. It can also be observed that the RSLoss has lower instability than NRSLoss, however, since NRSLoss eliminates the gradients from BN layer, the RSLoss actually gets lower instability by influencing BN layers, which in turn would hurt normal training, and thus hurt overall performance. The advantage of NRSLoss can also be interpreted using the NRSLoss landscape as shown in Figure \ref{nrsloss}. Compared to RSLoss, NRSLoss takes account of the channel importance, so that using NRSLoss can avoid regularizing neurons that are in the important channels. From these results, we can again conclude that neuron stability is important for certified robustness, in particular, IMP+NRSLoss motivated by improving neuron stability is effective for improving certified robustness.  

 We note that the literature results as in \cite{xiao2018training} are much better than the results reported in Table \ref{tb_2.255} and \ref{tb_8.255}. In fact, this is because they use much larger networks, whereas we focus on complete verification which causes Out-Of-Memory error on large networks, so we only choose small networks for testing, and note that our designed network, though small, is still the largest network we can perform complete verification on a GPU card with 24GB memory.

\subsubsection{Existence of certified lottery tickets.}

As one last ``hidden gem" finding, we demonstrate the existence of \textit{certified lottery tickets}, that generalizes the lottery ticket hypothesis \cite{frankle2018lottery} to certified robustness. Specifically, from Table \ref{tb_2.255}, we observe that {\bf all} pruning methods under certified training across all 3 datasets can find certified lottery tickets that can match {\it both} standard and verified accuracies to the original dense models, and most of the pruning methods can produce certified lottery tickets that significantly outperform original dense networks. From Table \ref{tb_8.255}, we see that certified lottery tickets can be found on most pruning methods with a bigger perturbation scale except for random pruning and IMP+RSLoss. From Figure \ref{training curves}(a), we observe that except for unstructured pruning (except for random pruning), certified lottery tickets occur almost in every sparsity. The above findings hence validate the existence of certified lottery tickets.

% \begin{figure*}[ht]
% \centering
%      \begin{subfigure}{.3\linewidth}
%          \includegraphics[width=1.1\linewidth]{lirpa_unstruct.pdf}
%          \caption{auto-LiRPA unstructured}
%     \label{Time}
%      \end{subfigure}
%      \hfill
%      \begin{subfigure}{.3\linewidth}
%          \includegraphics[width=1.1\linewidth]{lirpa_struct.pdf}
%          \caption{auto-LiRPA structured}
%     \label{Memory}
%      \end{subfigure}
%      \hfill
%      \begin{subfigure}{.3\linewidth}
%          \includegraphics[width=1.1\linewidth]{fgsm_unstruct.pdf}
%          \caption{FGSM unstructured}
%     \label{Memory}
%      \end{subfigure}
%         \caption{PlaceHolder: FashionMNIST dataset curves.}
%          %\vspace{-0.3em}
%         \label{training curves}
% \end{figure*}

% \begin{figure*}[ht]
% \centering
%      \begin{subfigure}{.3\linewidth}
%          \includegraphics[width=1.1\linewidth]{lirpa_unstruct.pdf}
%          \caption{auto-LiRPA unstructured}
%     \label{Time}
%      \end{subfigure}
%      \hfill
%      \begin{subfigure}{.3\linewidth}
%          \includegraphics[width=1.1\linewidth]{lirpa_struct.pdf}
%          \caption{auto-LiRPA structured}
%     \label{Memory}
%      \end{subfigure}
%      \hfill
%      \begin{subfigure}{.3\linewidth}
%          \includegraphics[width=1.1\linewidth]{fgsm_unstruct.pdf}
%          \caption{FGSM unstructured}
%     \label{Memory}
%      \end{subfigure}
%         \caption{PlaceHolder: SVHN dataset curves.}
%          %\vspace{-0.3em}
%         \label{training curves}
% \end{figure*}

\subsubsection{How should we choose pruning methods for certified robustness?}

Generally, we would recommend IMP and IMP+NRSLoss for performance concerns because they have the best verified accuracy under certified training across different datasets, and we recommend Network Slimming for efficiency concerns because its structured pruning nature can essentially reduce the computational overhead of the complete verification. We empirically find that the relative performance of our tested pruning methods under certified training is similar to that under standard training. We conjecture that this is because an important goal of most pruning methods is causing a minimal negative influence on the training objective function, and this objective function is benign accuracy under standard training and verified accuracy under certified training, respectively. These pruning methods also implicitly regularize the network stability and bound tightness as stated in Section \ref{benefits}, which leads to general improvement compared to dense baselines. However, our proposed NRSLoss-based pruning explicitly regularizes network stability which makes it outperform other pruning methods.

\subsection{Ablation}

In this section, we conduct several ablation studies mainly on the CIFAR10 dataset to further consolidate our claims. 

\subsubsection{Why focusing on complete verification}

In this subsection, we show the reasons why we focus on complete verification instead of both complete verification and incomplete verification. Firstly, we show that the bound produced by complete verification methods is always tighter than incomplete verification methods such as IBP or CROWN. Secondly, complete verification is further needed because neuron stability is important for the sub-domain split problem of complete verification, which is one motivation for proposing NRSLoss. To verify the first point, we test certified robustness under certified training about the comparison of incomplete/complete verification and suggest comparing Table \ref{tb_incomp} and Figure \ref{training curves}(c) for adversarial training setting, these results demonstrate that complete verification ($\beta$-CROWN) is always tighter than incomplete verification (IBP, CROWN).

\subsubsection{Results under different evaluation criterion}
In Section \ref{training methods}, we mentioned that the final numerical results are the averaged result of 5 different random seeds of the same training and pruning iteration. We here present another evaluation criterion, i.e. by first picking the iteration with the best verified accuracy, and then averaging the results of the picked iterations of the 5 different random seeds. The comparison of these 2 criteria is shown in Table \ref{tb_incomp}, where the ``AVG.'' column denotes the first criterion, while the ``BEST'' column denotes the second criterion. We observe slightly higher results under the second criterion, but the relative performance among different pruning methods is similar of these two criteria.

\begin{figure*}[!h]
\centering
     \begin{subfigure}{.4\linewidth}
         \includegraphics[width=1\linewidth]{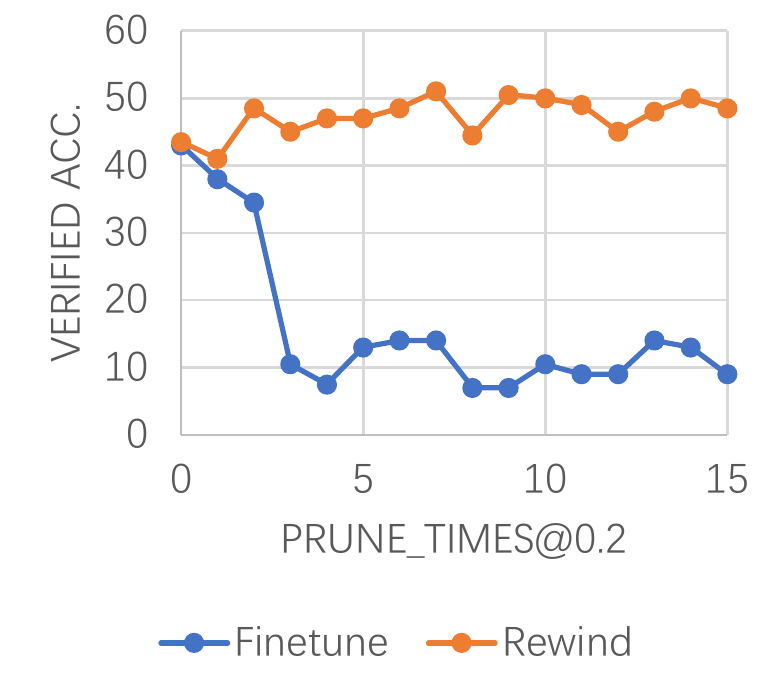}
         \caption{IMP}
             \label{Time}

     \end{subfigure}
     \begin{subfigure}{.4\linewidth}
         \includegraphics[width=1\linewidth]{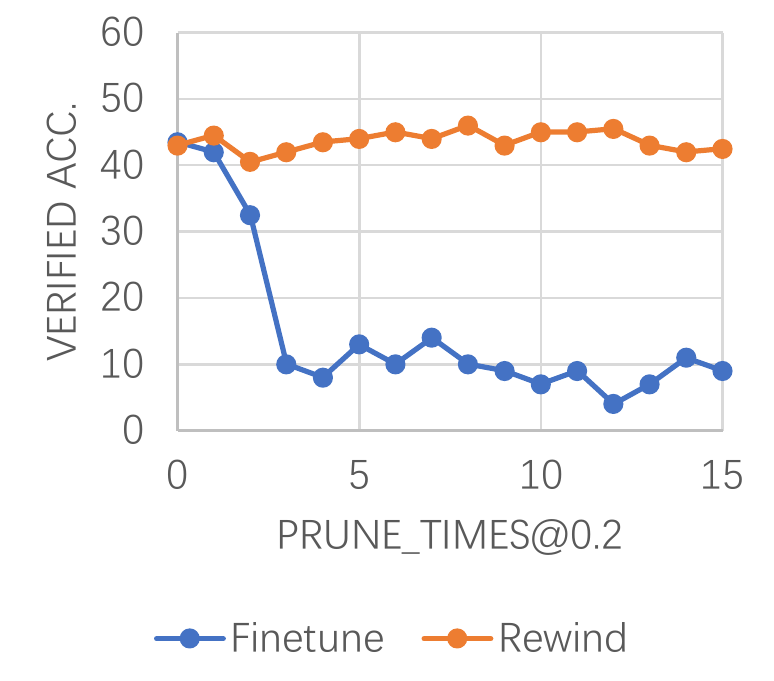}
         \caption{IMP+NRSLoss}
             \label{Memory}

     \end{subfigure}
        \caption{Comparison of finetuning and weight-rewinding  verified accuracy v.s. pruning times under auto-LiRPA setting.}
        \label{finetune}
\end{figure*}

\subsubsection{Comparison of pruning under different certified training methods}
In our main experiments, we choose the auto-LiRPA as the certified training method. The reason we choose this method is based on its training efficiency and competitive performance, and its training efficiency mainly comes from the {\it loss fusion} technique as proposed in \cite{xu2020automatic}. The training efficiency is important in our experiments because we use iterative training and pruning, which boosts the overall training time to 16 times longer. We notice that the certified training method\cite{shi2021fast} (denoted as {\bf FastIBP} in the following context) with SOTA performance (i.e. SOTA verified accuracy) claims that loss-fusion has a negative influence on the performance and thus doesn't adopt it in their method. We empirically find that without loss-fusion, the training speed of FastIBP is 4 times slower than auto-LiRPA. We thus choose auto-LiRPA as the certified training method in our main experiments. However, we here present a comparison of results (see Table \ref{tb_fastibp}) of these 2 certified training methods on the CIFAR10 dataset and pruned with several pruning methods, to demonstrate that the improvement of certified robustness brought by pruning is consistent with different certified training methods. The hyperparameter settings are the same as mentioned in our main experiments. From Table \ref{tb_fastibp}, we observe better performance can be obtained with FastIBP, and standard/verified accuracies are consistently improved with different pruning methods, among which IMP+NRSLoss still achieves the best performance.

\begin{table*}[h]
\centering
%\vspace{-0.5em}
\caption{Comparison of auto-LiRPA and Fast-IBP on CIFAR10 dataset}
\label{tb_fastibp}
\begin{center}
\begin{small}
\begin{sc}
\begin{tabular}{c|c|c|c c c| c| c c  c }
\toprule
\multicolumn{3}{c|}{Training Method}&\multicolumn{3}{c|}{auto-LiRPA}&&\multicolumn{3}{c}{FastIBP} \\
\hline
\makecell{Pruning\\ type}& \makecell{Pruning\\ Method}& \makecell{Remain\\ Ratio}
&$std$ &$ver$ &$t$&\makecell{Remain\\ Ratio}
&$std$ &$ver$ &$t$
\\
\hline
\multicolumn{2}{c|}{Dense}&1 & 54.1 &43.0 &6.68 & 1 &56.0  & 43.5 & 8.42\\
\hline
%\multirowcell{5}{Unstru-\\ ctured}& Random& 0.4& 55.0&44.0& 5.25&0.26 &37.5  &31.0 &2.74   \\
& IMP&0.13 &61.0 &50.1 &6.65 & 0.04 &63.5 &53.0 & 6.17\\
& HYDRA&0.11 & 60.5 &48.3&8.75 &0.13 &64.0 &51.0 &6.56 \\
& IMP+RSLoss&0.13 &58.6  &46.3&6.52 &0.17 &61.5  &47.0 &5.98 \\
& IMP+NRSLoss&0.21 &62.2  &51.2&6.06 &0.17 &64.0 &54.0  & 5.94  \\

\hline
\multirowcell{1}{Structured}& Slim&0.79 &59.2  &47.5&5.65 &0.44 &56.0  &46.5 &5.53 \\
\bottomrule
\end{tabular}
\end{sc}
\end{small}
%\vspace{-0.5em}
\end{center}
\end{table*}

\begin{table}[h]
\centering
\caption{Comparison of verified accuracy of different incomplete verification and complete verification methods for models trained with auto-LiRPA and pruned with several representative pruning methods on CIFAR10.}
\label{tb_incomp}
\begin{center}
\begin{small}
\begin{sc}
\begin{tabular}{c|c|c|c|c}
\toprule
 &IBP & CROWN & \multicolumn{2}{c}{$\beta$-CROWN} \\
 \hline
  & & & avg. &best  \\
\hline
IMP &47.2 & 43.6 &50.1 &50.3 \\
HYDRA & 46.8 &43.1 & 48.3&49.3 \\
IMP+RS & 47.5 &42.1 & 46.3&48.6 \\
IMP+NRS & 47.3 & 43.4 & 51.2& 51.8 \\
SLIM &46.0  & 41.9 &47.5 & 47.8 \\
\bottomrule
\end{tabular}
\end{sc}
\end{small}
\end{center}
\end{table}

\subsubsection{Comparison of finetuning and weight-rewinding for pruning}
\label{finetune_rewind}

We empirically find that after each pruning, rewinding the network parameters to their initial states as in \cite{frankle2018lottery} produces better performance than finetuning the parameters as in \cite{sehwag2020hydra}. Specifically, we follow the experiment setup as in Section \ref{setup}, except that for finetuning mode we don't re-initialize the learning rate after pruning. The results are demonstrated in Figure \ref{finetune}. We observe that the finetuning-based pruning always produces worse performance than weight rewinding-based pruning, and its accuracy tends to collapse at 3rd pruning iteration. Therefore we conclude that weight-rewinding-based pruning is more effective than finetuning-based pruning for certified robustness.

% \begin{figure*}[ht]
% \centering
%      \begin{subfigure}{.4\linewidth}
%          \includegraphics[width=1.\linewidth]{IMP_finetune.pdf}
%          \caption{IMP}
%     \label{Time}
%      \end{subfigure}
%      \hfill
%      \begin{subfigure}{.4\linewidth}
%          \includegraphics[width=1.\linewidth]{HYDRA_finetune.pdf}
%          \caption{HYDRA}
%     \label{Memory}
%      \end{subfigure}
%         \caption{Comparison of finetuning and weight-rewinding  verified accuracy v.s. pruning times under auto-LiRPA setting.}
%         \label{finetune}
% \end{figure*}

\subsubsection{The performance of HYDRA in original paper}
\label{hydra_lwm}

We reproduce the performance of HYDRA using the released code from \cite{sehwag2020hydra}. Specifically, we run their original experiments of HYDRA pruning and Least Weight Magnitude(LWM) pruning under the CROWN-IBP setting and SVHN dataset. The pruning is applied only once followed by finetuning. The results under different pruning rates $k$ are shown in Table \ref{tb_hydra}. We observe that LWM pruning produces better performance than HYDRA pruning. Since the pruning process of LWM is essentially the same as IMP in our experiments (except that IMP prunes multiple times instead of pruning only once), the result that LWM is better than HYDRA is consistent with our experiments in Table \ref{tb_2.255} where IMP is relatively better than HYDRA in many cases.

\begin{table}[h]
\centering
\caption{The verified accuracies under different sparsities of our reproduced CROWN-IBP experiments on SVHN dataset from \cite{sehwag2020hydra}. The verified accuracy is obtained using CROWN-IBP. The model is {\it CNN-large} as in \cite{sehwag2020hydra}. }
\label{tb_hydra}
\begin{center}
\begin{small}
\begin{sc}
\begin{tabular}{c|c|c}
\toprule
Remain Ratio & LWM & HYDRA \\
\hline
1 & \multicolumn{2}{c}{48.99} \\
\hline
0.1 &49.48 &48.60  \\
0.05  &45.56& 44.18 \\
0.01 &47.47 & 44.14 \\
\bottomrule
\end{tabular}
\end{sc}
\end{small}
\end{center}
\end{table}

\subsection{Summary of Findings}
In our experiments, we find that pruning can generally improve certified robustness for neural networks trained with different robust training methods and observe the existence of certified lottery tickets. Under adversarial training, we observe a significant trade-off between standard and verified accuracies with different pruning methods, but under certified training, pruning can improve both standard and verified accuracies. From our experiments, we know that RSLoss and NRSLoss are both effective at regularizing network stability but NRSLoss is better for imposing less regularization on more important neurons and removing the negative influence of stability regularization for BN layers. From Figure \ref{fig_resource}, we observe that structured pruning can considerably reduce the computational overhead of complete verification for neural networks. Among existing pruning methods that we tested, we empirically find that IMP can generally achieve relatively good performance, while it is outperformed by IMP+NRSLoss which incorporates stability regularization.

\section{Conclusion}
In this paper, we demonstrate that pruning can generally improve certified robustness, both for adversarial and certified training. We analyze some important factors that influence certified robustness, and offer a new angle to study the intriguing interaction between sparsity and robustness, i.e.
interpreting the interaction of sparsity and certified robustness
via neuron stability. In particular, we find neuron stability to be crucial for improving certified robustness, on which motivation we propose the novel NRSLoss-based pruning that outperforms existing pruning methods. We also observe the existence of certified lottery tickets. We believe our work has revealed the relationships between pruning and certified robustness, which can shed light on future research to design better sparse networks with certified robustness.





%\bibliographystyle{IEEEtranS}

\bibliographystyle{./IEEEtranS}
\bibliography{./IEEEabrv,./IEEEexample}

\end{document}